\documentclass{article}
\usepackage[preprint]{icml2026}

\usepackage{amsmath,amssymb,booktabs,array,graphicx}
\usepackage{placeins}
\usepackage{dblfloatfix}
\usepackage{flafter}
\usepackage[hidelinks]{hyperref}
\icmltitlerunning{Hidden Sensitivity in LLM Pre-training}

\begin{document}
\twocolumn[
\icmltitle{Hidden Sensitivity in LLM Pre-training: A Dynamic System View}
\begin{icmlauthorlist}
\icmlauthor{Bo Liu}{cuhk}
\icmlauthor{Yuan Xie}{cuhk}
\icmlauthor{Junan Chen}{cuhk}
\icmlauthor{Fujun Han}{cuhk}
\end{icmlauthorlist}
\icmlaffiliation{cuhk}{The Chinese University of Hong Kong, Shenzhen, China}
\icmlcorrespondingauthor{Fujun Han}{hanfujun@cuhk.edu.cn}
\icmlkeywords{optimization, language models, training dynamics, perturbation sensitivity}
\vskip 0.3in
]
\printAffiliationsAndNotice{}
\begin{abstract}
In large language model (LLM) pre-training, the loss curve can conceal substantial instability: even small perturbations can produce dramatically different predictions while leaving the loss nearly unchanged. To study this phenomenon and its underlying mechanisms, we formulate LLM pre-training as a non-autonomous dynamical system that jointly evolves model weights and Adam optimizer states, and use this formulation to define a normalized sensitivity score in optimizer-relative coordinates that identifies sensitive training checkpoints. At one such checkpoint, a weight perturbation of only $10^{-5}$ changes 35.7\% of top-1 predictions and 16.2\% of benchmark answers after 200 training steps, and only 0.8\% of greedy generations remain identical, although validation loss stays nearly unchanged. Perturbations to Adam's first and second moments have comparably large effects, and the same pattern appears in OLMo-1B. Stabilization timed by this diagnostic reduces top-1 disagreement to 0.83\% and raises generation exact match to 88.3\%, at the cost of a 0.11-nat increase in validation loss. These results expose an optimizer-state-dependent sensitivity that loss-based evaluation overlooks, and show that it can be diagnosed and mitigated.
\end{abstract}

\begin{figure}[!t]
\centering
\includegraphics[width=\columnwidth]{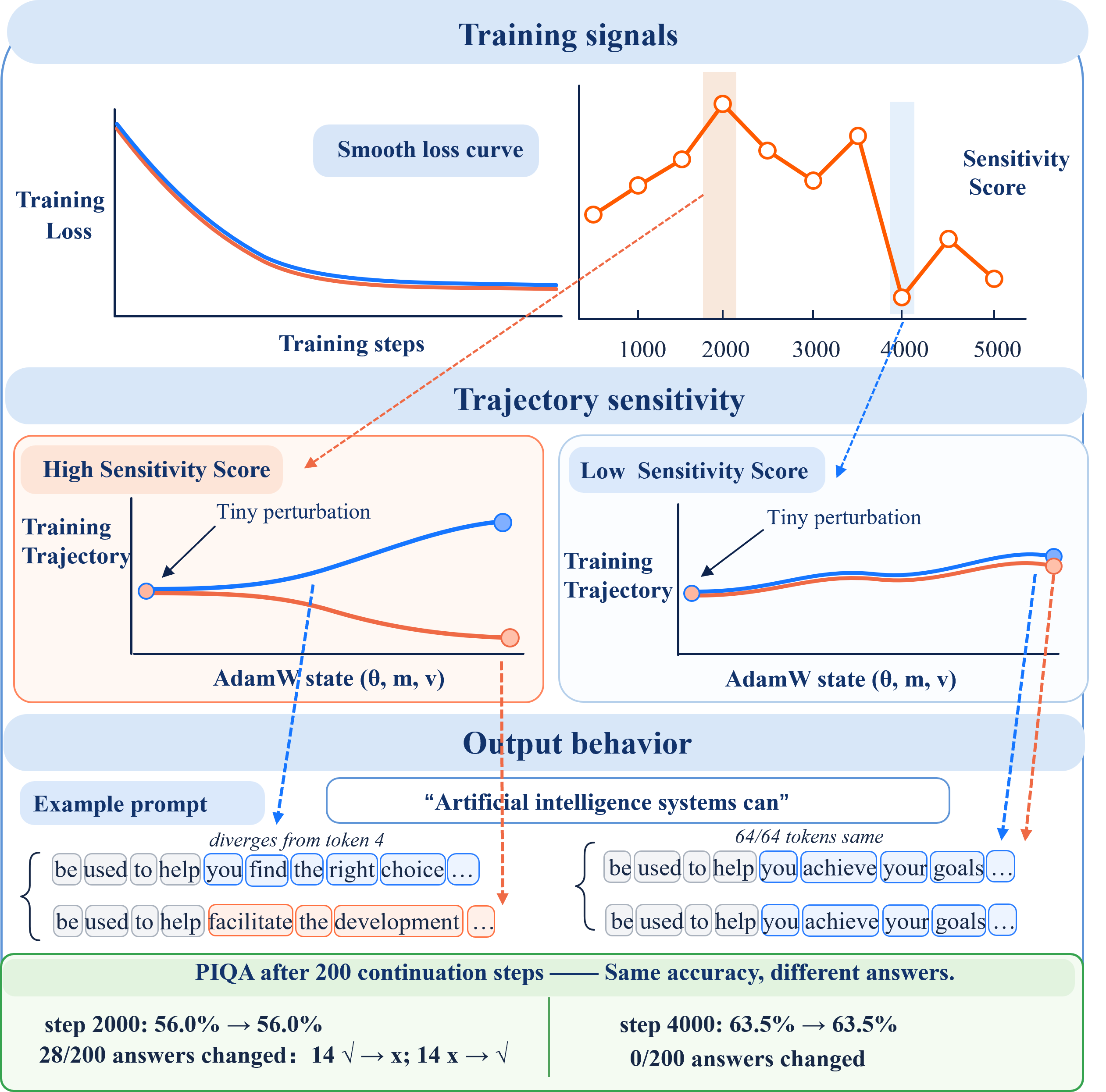}
\caption{Loss is smooth while sensitivity varies sevenfold: a $10^{-5}$ perturbation changes 28 of 200 PIQA answers at unchanged 56.0\% accuracy from step 2000, but none from step 4000.}
\label{fig:teaser}
\end{figure}

\section{Introduction}
Small perturbations are everywhere: batch order, dropout masks, numerical kernels, and optimizer state introduce differences that cannot be eliminated. The Butterfly Effect for pre-training is when such a difference is amplified into a different model. Loss and benchmark accuracy can look stable while two runs answer different questions correctly \citep{summers2021nondeterminism,damour2022underspecification}. Aggregate metrics compare average performance, not whether training preserves the state from which models emerge. We locate the training windows where a small state difference is amplified and follow it from optimizer state to updates, parameters, logits, and individual predictions.

Studies of Transformer instability examine loss divergence, gradient scale, and adaptive updates \citep{molybog2023theory,wortsman2024smallscale,huang2025spam}. Floating-point computations can introduce discrepancies even when the intended training procedure is unchanged \citep{shanmugavelu2024floating}. Sensitivity asks what happens if the checkpoint is slightly changed and training continues: an update can be small in the original run yet change substantially under that perturbation.

We study this response in AdamW language-model training \citep{kingma2015adam,loshchilov2019decoupled}. Because AdamW carries gradient history in its first and second moments, weights alone do not determine the next iterate; we include this memory in the state $(\theta,m,v)$ and differentiate the full update. Dynamical analyses of learning motivate measuring trajectory separation \citep{li2017stochastic,herrmann2022chaotic}, and finite-time Lyapunov exponents measure displacement growth over a window, depending on that window, the metric, and the initial direction \citep{sharafi2025weight}. We build a related finite-window score in relative coordinates tied to weights, preconditioned updates, and adaptive scaling. A raw Euclidean norm mixes parameters, gradients, and squared gradients in incompatible units; our relative coordinates keep them on a common scale, so moment differences enter alongside weight differences.

Our central test perturbs one checkpoint component and continues both branches on the same future batches, isolating state amplification and connecting the local diagnostic to later predictions. Sensitivity varies during smooth optimization, moment perturbations affect updates as strongly as weight perturbations, and after 1000 steps top-1 disagreement reaches 27\% from the sensitive checkpoint versus 1--1.5\% from an insensitive one. Independent seeds and an OLMo-1B continuation reproduce the pattern.

Stabilization must balance trajectory agreement against progress: stronger damping lowers separation to about 0.14 but raises validation loss to 4.143. At matched strength and activation, gradient-norm, update-norm, and sensitivity schedules differ only in placement, and the sensitivity timing cuts separation a further 7\% at comparable loss.

Our contributions are:
\begin{enumerate}
    \item \textbf{A striking phenomenon.} We find that a tiny perturbation
    to one checkpoint changes a large share of top-1 predictions after a
    short continuation, while validation loss and aggregate accuracy stay
    close. The amplification concentrates in short training windows and
    recurs across seeds and model size.
    \item \textbf{A diagnostic metric.} We formulate LLM pre-training as
    a dynamical system and define a sensitivity score that measures
    displacement in weights, update directions, and preconditioner state on
    a common scale. High-scoring checkpoints produce far more top-1
    disagreement over the long term than low-scoring ones.
    \item \textbf{An effective control.} We schedule AdamW stabilization
    where the offline sensitivity scores are highest. With the same damping
    budget, this timing sharply reduces top-1 disagreement and recovers
    almost all greedy generations at a small validation-loss cost.
\end{enumerate}

\section{Related Work}
\paragraph{Sensitivity of optimization.}
Algorithmic stability studies how training-data changes affect the learned predictor \citep{hardt2016train}. For a fixed objective, edge-of-stability analysis relates gradient descent to loss curvature \citep{cohen2021gradient}; dynamical stability constrains the minima selected by SGD \citep{wu2023implicit,chemnitz2025stability}. Adaptive optimizers add memory-dependent effects: Adam's preconditioner can trigger loss spikes \citep{bai2025adaptive}, while cautious optimizers modify updates according to momentum--gradient agreement \citep{liang2026cautious}. Open checkpoints support analysis across training stages \citep{biderman2023pythia}. We perturb checkpoint weights or moments while replaying identical future batches, then measure trajectory separation and prediction disagreement during otherwise smooth optimization.

\paragraph{Dynamical systems in deep learning.}
Weinan E proposed representing high-dimensional functions through continuous dynamical systems \citep{e2017proposal}. Subsequent work derives training algorithms from Pontryagin's maximum principle \citep{li2018maximum}, designs stable architectures through differential-equation analysis \citep{haber2017stable}, and learns continuous hidden-state dynamics with neural ODEs \citep{chen2018neuralode}. Here network depth acts as evolution time. A complementary approach treats optimization steps as time: stochastic gradient algorithms are approximated by stochastic differential equations \citep{li2019modified}, and transient-chaos analysis examines learning at large learning rates \citep{jimenez2026leveraging}. FTLE also measures input sensitivity across network depth \citep{storm2024finite}. Our analysis follows perturbations across training steps, with weights and Adam memory forming the evolving state; paired continuations connect local sensitivity to later behavior.

\section{Method}
\subsection{Training Dynamics}
Let $B_t$ be the batch at step $t$ and $g_t=\nabla_\theta\ell(\theta_t;B_t)$. AdamW remembers past gradients in two moving averages: $m_t$ tracks their direction, while $v_t$ tracks their squared magnitude and sets a separate scale for each coordinate. With learning rate $\eta_t$ and weight decay $\lambda_w$, its updates are
\begin{align}
m_{t+1}&=\beta_1m_t+(1-\beta_1)g_t,\label{eq:adam-m}\\
v_{t+1}&=\beta_2v_t+(1-\beta_2)g_t^2,\label{eq:adam-v}\\
u_{t+1}&=\frac{\widehat m_{t+1}}{\sqrt{\widehat v_{t+1}}+\epsilon},\label{eq:u}\\
\theta_{t+1}&=(1-\eta_t\lambda_w)\theta_t-\eta_tu_{t+1}.
\end{align}
Products, squares, and divisions of vectors are elementwise; hats denote bias correction. In particular, $u_{t+1}$ is the Adam direction before multiplying by the learning rate and excludes decoupled weight decay.

Two checkpoints with identical weights can take different next steps if their moment memories differ. We therefore define the training state as $x_t=(\theta_t,m_t,v_t)$. Given the batch and optimizer settings, this state determines $x_{t+1}=T_t(x_t)$. The map changes with the batch and learning-rate schedule. Replaying the same batches and schedule in both branches isolates the amplification of their initial state difference; changing batch order changes the map itself.

To follow a very small difference between states, write its first-order approximation as $q_t=(q_{\theta,t},q_{m,t},q_{v,t})$. The corresponding gradient change is $h_t=H_tq_{\theta,t}$, where $H_t$ is the batch-loss Hessian. Differentiating the update gives
\begin{align}
q_{m,t+1}&=\beta_1q_{m,t}+(1-\beta_1)h_t,\label{eq:q-m}\\
q_{v,t+1}&=\beta_2q_{v,t}+2(1-\beta_2)g_t\odot h_t,\label{eq:q-v}\\
q_{u,t+1}&=\frac{q_{\widehat m,t+1}}{d_t}
-\frac{\widehat m_{t+1}\odot q_{\widehat v,t+1}}
{2\sqrt{\widehat v_{t+1}}\odot d_t^2},\label{eq:tangent}\\
q_{\theta,t+1}&=(1-\eta_t\lambda_w)q_{\theta,t}-\eta_tq_{u,t+1},\label{eq:q-theta}
\end{align}
where $d_t=\sqrt{\widehat v_{t+1}}+\epsilon$. A weight difference first changes the gradient through $h_t$. That gradient difference enters both moment memories, changes the update direction, and produces the next weight difference. The recurrence explains why a perturbation to optimizer memory can influence the model even when its initial weights are unchanged. Hessian--vector products evaluate this interaction without materializing the Hessian. Appendix~\ref{app:implementation} describes how the differentiable AdamW map relates to clipping, rounding, and quantized state.

\subsection{Finite-Time Sensitivity}
For $J_t=D T_t(x_t)$, a conventional directional finite-time exponent is
\begin{equation}
\lambda_{t,K}(q)=\frac1K\log\frac{
\|J_{t+K-1}\cdots J_tq\|_{t+K}}{\|q\|_t}.
\label{eq:ftle}
\end{equation}
The numerator measures how large the initial displacement becomes after the window, and the logarithm gives its average multiplicative growth per step: positive values indicate amplification and negative values contraction in the chosen distance. The exponent depends on window, metric, and direction. Maximizing over directions gives the maximal FTLE. Our probe instead follows a Gaussian parameter-block direction through the full state and reports a renormalized finite-window score (Appendix~\ref{app:implementation}).

\subsection{Relative State Geometry}
The tangent recurrence identifies what propagates, but comparing its components requires a choice of distance. A raw Euclidean norm adds perturbations to weights, gradients, and squared gradients, so its value depends on their units: rescaling the loss by $c$ rescales $m$ by $c$ and $v$ by $c^2$ while leaving the Adam direction unchanged away from the numerical floor.

We measure relative weight displacement, relative update displacement, and multiplicative change to the preconditioner. The dynamical state remains $(\theta,m,v)$: the update direction $u$ and log-preconditioner $\rho$ are functions of the moments. They provide coordinates for measuring a state displacement. For tensors indexed by $\ell=1,\ldots,L$, define
\begin{equation}
\mathcal R(a;b)^2=\frac1L\sum_\ell
\frac{\operatorname{RMS}(a_\ell)^2}{\operatorname{RMS}(b_\ell)^2+f^2}.
\end{equation}
Here RMS is the square root of the mean squared entries of a tensor. Each ratio compares a displacement with the typical size of its corresponding tensor, rather than dividing by individual entries that may be close to zero. The floor is $f=10^{-12}$. Equal tensor averaging prevents a large embedding tensor from dominating by size alone. The denominator of the Adam update controls how strongly each gradient coordinate affects the weights. We use its logarithm, $\rho=\log(\sqrt{\widehat v}+\epsilon)$, so that a change measures the fractional change in this scaling. Its tangent is
\begin{equation}
q_\rho=\frac{q_{\widehat v}}{2\sqrt{\widehat v}\odot(\sqrt{\widehat v}+\epsilon)}.
\end{equation}
The norm used to evaluate each local tangent is
\begin{equation}
N(q)^2=\mathcal R(q_\theta;\theta)^2+
\mathcal R(q_u;u)^2+\frac1L\sum_\ell\operatorname{RMS}(q_{\rho,\ell})^2.
\label{eq:norm}
\end{equation}
The first term measures how far the weights move relative to their own size. The second measures how much the direction of the next update changes, and the third measures how much its adaptive scaling changes. We give these three contributions equal weight. Finite-time growth depends on the metric; fixing it gives a common basis for checkpoint comparison. Appendices~\ref{app:geometry} and \ref{app:derivation} give the metric transformation and full Jacobian.

\subsection{Checkpoint Estimation}
The probe starts from $(q_\theta,0,0)$, propagates this direction through the Adam tangent recurrence, and evaluates Equation~\ref{eq:norm}. It uses three local iterations on a fixed checkpoint batch for the small model and one for OLMo, leaving the checkpoint unchanged. Algorithm~\ref{alg:probe} in Appendix~\ref{app:implementation} specifies the normalization and the reported score. We test the checkpoint ranking with separate paired continuations on subsequent shared batches.

For a twice-differentiable realized $K$-step map $F$,
\begin{equation}
F(x+\delta)-F(x)=DF(x)\delta+O(\|\delta\|^2).
\end{equation}
The linear term describes the response while the displacement is small; as the branches separate, nonlinear effects change it, so we use actual paired continuations to measure later separation and output differences rather than extrapolating the local tangent through training.

\section{Experiments}
\subsection{Experimental Setup}
We use a LLaMA-style model with 76.8M implemented parameters, previously labeled ``60M'' in run names, trained on C4 \citep{raffel2020exploring} for 5000 steps, and continue OLMo-2-0425-1B (1.485B parameters) \citep{groeneveld2024olmo} on Dolma \citep{soldaini2024dolma} for 2000 steps. The evidence comprises 364 paired trajectories, summarized in Table~\ref{tab:coverage}; Appendix~\ref{app:settings} lists settings and grouping.

At each checkpoint, one branch is unchanged and another receives a perturbation to exactly one of $\theta$, $m$, or $v$; both then replay the same batches. Nominal magnitudes are $10^{-6}$, $10^{-5}$, and $10^{-4}$ for the small model, with the largest omitted for OLMo, and second moments are clamped to remain nonnegative. Separate conditions change batch order. Short continuations last 200 steps; selected checkpoints are followed for 1000 or 2000 steps. Repeated amplitudes and channels at one checkpoint are dependent conditions, so we report checkpoint-level results.

We measure relative separation $\|z^{\rm pert}-z^{\rm ref}\|_2/(\|z^{\rm ref}\|_2+10^{-12})$ in updates, parameters, and gradients, and use symmetric logits KL, top-1 disagreement, and greedy-generation exact match for outputs. Multiple-choice disagreement measures changed answers separately from net accuracy on PIQA, HellaSwag, and ARC-Easy \citep{bisk2020piqa,zellers2019hellaswag,clark2018arc}, which test physical commonsense, continuation selection, and elementary science. These readouts show whether comparable aggregate performance also means comparable predictions on individual inputs.

\begin{figure*}[t]
\centering\includegraphics[width=\textwidth]{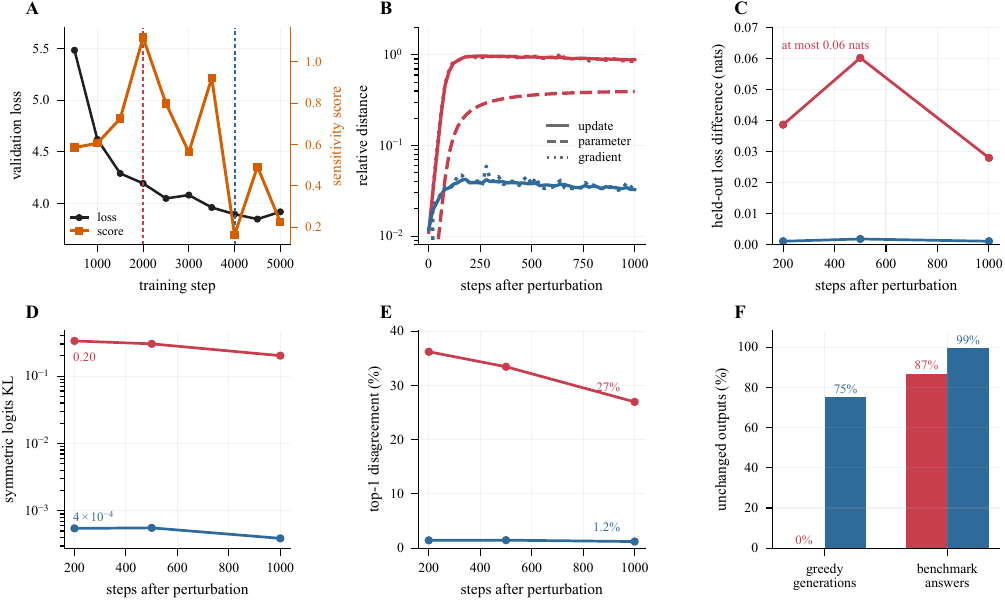}
\caption{One perturbation, two checkpoints. Red: sensitive checkpoint (step 2000); blue: stable checkpoint (step 4000). A: validation loss and sensitivity score. B--F: update, parameter, and gradient separation; held-out loss; logits KL; top-1 disagreement; unchanged outputs after a $10^{-5}$ weight perturbation.}
\label{fig:phenomenon}
\end{figure*}

\subsection{Checkpoint Sensitivity}
\begin{figure*}[t]
\centering
\includegraphics[width=\textwidth]{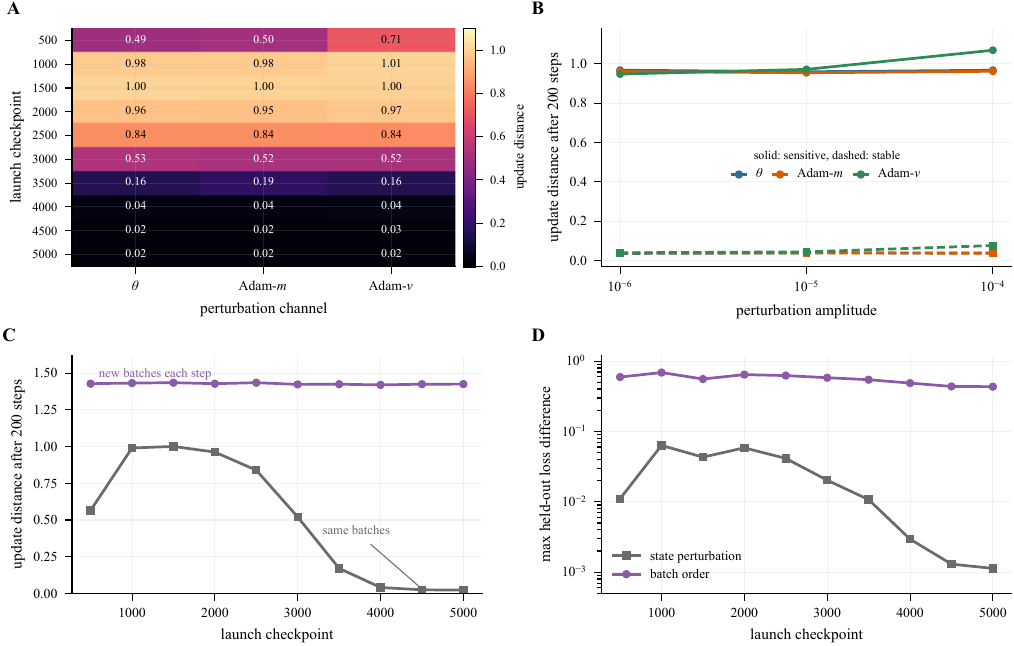}
\caption{Checkpoint and channel landscape. A: 200-step update separation at $10^{-5}$ across ten checkpoints and three state channels. B: separation versus amplitude; solid/dashed: sensitive/stable checkpoint. C--D: batch-order versus state perturbations for update separation and held-out loss (log scale).}
\label{fig:landscape}
\end{figure*}

\paragraph{Checkpoint variation.}
Validation loss decreases from 5.48 to 3.92 on the dense small-model trajectory, while the normalized diagnostic spans 0.163 at step 4000 to 1.116 at step 2000 (Figure~\ref{fig:phenomenon}A). Figure~\ref{fig:landscape}A tests the response at all ten sampled checkpoints. Each checkpoint receives the same channel and amplitude conditions, followed by the same continuation horizon; separation changes across the matrix for both updates and parameters, so the response depends on where the perturbation enters training.

At nominal magnitude $10^{-5}$, weight perturbations produce 200-step relative update distances of 0.960 at step 2000 and 0.039 at step 4000. At the sensitive checkpoint the perturbed update differs from the reference by almost its full norm, so the branches take nearly independent steps from an imperceptible initial difference; at the stable checkpoint the updates agree to within four percent of their norm, so they continue along nearly the same trajectory. Both values lie on opposite sides of the 90-condition mean of 0.505, a change in regime rather than a small modulation around a common level. Because the two branches replay the same batches, the disparity can only come from how the state amplifies the initial difference.

\paragraph{Perturbation response.}
The same pattern appears when the perturbation enters through Adam memory: across 90 state-perturbation conditions, both moment channels follow the checkpoint pattern of weight perturbations (Figure~\ref{fig:landscape}A). Appendix Figure~\ref{fig:perturbation_matrix} gives the full batch-order and loss comparisons. At $10^{-5}$, mean 200-step update distances are 0.505 for weights, 0.507 for first moments, and 0.529 for second moments. Comparable responses across all three channels mean that a checkpoint's sensitivity cannot be read from its weights alone.

The channels differ in how their response depends on amplitude: second-moment mean parameter separation rises from 0.113 at $10^{-6}$ to 0.171 at $10^{-5}$ and 0.818 at $10^{-4}$, while weight and first-moment separation changes little, and even second-moment 200-step update separation stays nearly flat (Figure~\ref{fig:landscape}B). This flattening marks saturation, so we measure growth during the continuation as well as separation at its endpoint.

Batch-order changes give a different comparison. Their mean update distance is near 1.43, consistent with the continuing effect of different input streams, and held-out loss differences stay below 0.06 nats for state perturbations and near 0.5 nats for batch-order changes. Unlike an initial state perturbation, this condition repeatedly injects differences into the branches, so we analyze it separately.

\subsection{Behavioral Effects}
\paragraph{Long-term effects.}
We follow the selected high- and low-sensitivity checkpoints for longer and evaluate their outputs on shared inputs; Figure~\ref{fig:phenomenon} shows the response. At 1000 continuation steps, perturbations from the high-sensitivity checkpoint reach relative parameter distances of 0.39--0.40 and gradient distances near 0.85. From the low-sensitivity checkpoint, the corresponding distances remain below 0.015 and near 0.03. This contrast persists to 2000 steps across weights and both moment channels; Appendix Figure~\ref{fig:long_window} gives the full trajectories.

Output differences follow the same ordering. High-checkpoint continuations have logits KL near 0.20 and about 27\% top-1 disagreement; low-checkpoint continuations have KL near $4$--$6\times10^{-4}$ and 1--1.5\% disagreement. Average held-out loss differs by at most 0.06 nats, even as the branches disagree on which token to choose. At 1000 steps, 75\% of greedy generations from the low-sensitivity checkpoint remain exact versus 0\% from the high-sensitivity one, and 99\% versus 87\% of benchmark answers are unchanged.

Low-margin predictions explain how this happens. Average cross-entropy weights all positions equally, so changing the leading token where the model is nearly undecided moves the mean loss only slightly, just as individually flipped benchmark answers can cancel when gains on some examples offset losses on others. Net loss and accuracy therefore stay near their reference values even when a sizeable share of predictions differs. Greedy generation compounds the effect instead of averaging it: one early flip changes the prefix, and every later token is conditioned on that prefix, so exact match can fall from most generations to none while the per-position distribution changes far less. The three readouts measure different objects, mean continuous fit, per-decision agreement, and whole-sequence agreement, so a run judged by loss and aggregate accuracy alone can look like a valid continuation after its predictions and generations have mostly changed.

\begin{figure*}[!tb]
\centering
\includegraphics[width=\textwidth]{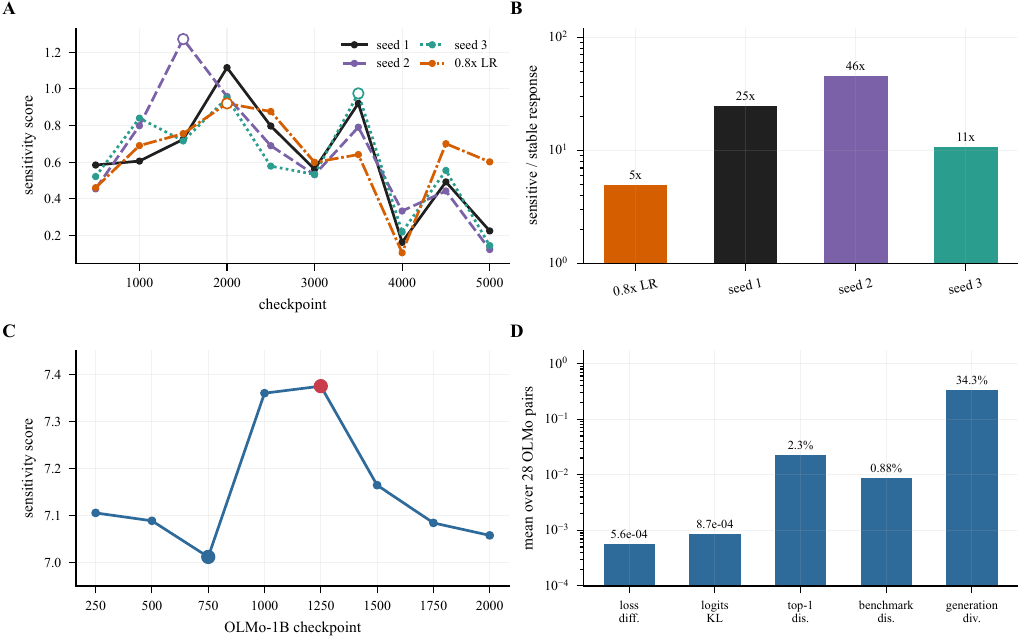}
\caption{Sensitivity transfers across runs. A: sensitivity scores across checkpoints for three seeds and a $0.8\times$-learning-rate run; hollow markers: selected sensitive checkpoints. B: high-to-low response ratio. C--D: OLMo-1B scores and output change after 1000 steps.}
\label{fig:transfer}
\end{figure*}

\paragraph{Seed replication.}
The high/low response contrast recurs across three independent small-model seeds, with different timing and magnitude, as Table~\ref{tab:seeds} and Figure~\ref{fig:transfer}A summarize. High-to-low response ratios range from $5\times$ to $46\times$, as Figure~\ref{fig:transfer}B shows. Table~\ref{tab:seeds} lists the checkpoint pair behind each row: the high checkpoint falls between steps 1500 and 3500 and the low one between 4000 and 5000, so the contrast does not follow a fixed schedule. A $0.8\times$-learning-rate AdamW run retains a 0.783/0.160 high/low update-distance contrast, so the effect does not depend on optimization speed. Repeated tangent seeds give similar diagnostic values conditional on a batch, while changing the diagnostic batch produces greater variation. Appendix Figure~\ref{fig:replication_full} reports these controls and the independent runs.
\begin{table}[t]
\caption{Seed replication. Response is median update separation across perturbation channels at each run's high- and low-score checkpoint.}
\label{tab:seeds}\centering
\resizebox{\linewidth}{!}{%
\begin{tabular}{lrrrr}\toprule
Seed & High step & Low step & High response & Low response\\\midrule
1 & 2000 & 4000 & 0.960 & 0.039\\
2 & 1500 & 5000 & 0.957 & 0.021\\
3 & 3500 & 5000 & 0.236 & 0.022\\\bottomrule
\end{tabular}}
\end{table}

\paragraph{OLMo-1B results.}
OLMo provides one continuation trajectory with eight diagnostic checkpoints and 28 perturbation conditions evaluated at both 200 and 1000 steps. It differs in corpus, tokenizer, and recipe, so a shared response does not depend on one pipeline. Table~\ref{tab:olmo_main} reports loss, logits KL, top-1, generation, and benchmark readouts. The loss difference shrinks by half between 200 and 1000 steps while top-1 disagreement and generation match barely move, so the horizon changes the continuous distance more than the discrete decisions. Mean absolute accuracy changes remain below 0.75 percentage points. The OLMo scores select steps 750 and 1250, as Figure~\ref{fig:transfer}C shows; continuations from these checkpoints reproduce the small-model pattern at 1.485B parameters (Figure~\ref{fig:transfer}D), and Table~\ref{tab:olmo_main} shows the same split across channels. Appendix Figures~\ref{fig:olmo_transfer} and \ref{fig:olmo_horizon} give channel and horizon breakdowns.

\begin{table}[t]
\caption{OLMo-1B output comparison at 200 and 1000 continuation steps.}
\label{tab:olmo_main}\centering
\small
\begin{tabular}{lrr}\toprule
Observable & 200 steps & 1000 steps\\\midrule
Absolute loss difference & $5.61\!\times\!10^{-4}$ & $2.30\!\times\!10^{-4}$\\
Symmetric logits KL & $8.75\!\times\!10^{-4}$ & $8.32\!\times\!10^{-4}$\\
Top-1 disagreement & 2.27\% & 2.30\%\\
Generation exact match & 65.7\% & 63.6\%\\
PIQA disagreement & 0.23\% & 0.45\%\\
HellaSwag disagreement & 1.48\% & 1.25\%\\
ARC-Easy disagreement & 0.91\% & 1.02\%\\\bottomrule
\end{tabular}
\end{table}

\paragraph{Muon/AdamW hybrid.}
The 1.485B model also runs under a Muon/AdamW hybrid, with Muon on the two-dimensional hidden matrices and AdamW on the embeddings, head, and vectors, and it shows the same ordering. Muon orthogonalizes the momentum update instead of preconditioning it with a second moment, so its state carries sensitivity differently from AdamW. Averaged over six conditions per checkpoint, relative parameter separation contracts from $2.24\times10^{-4}$ to $3.60\times10^{-5}$, while symmetric logits KL grows from $6.5\times10^{-4}$ to $1.0\times10^{-3}$ and generation agreement falls from 59.4\% to 39.6\%. Answer flips summed over the six conditions rise from 2 to 14 (192 questions each), while top-1 disagreement stays near 1.3\%.

Perturbing the Muon momentum buffer is as consequential as perturbing the weights: across nine matched amplitudes the median momentum-to-weight ratio is 0.97 for update separation and 0.93 for parameter separation. Because the buffer is orthogonalized before each update, perturbing it changes the direction of later updates even when the weights are identical. Within the state perturbations, the second moment carries the amplitude response: raising $\epsilon$ from $10^{-6}$ to $10^{-4}$ lifts parameter separation from $4.1\times10^{-4}$ to $1.8\times10^{-3}$, while the other channels stay near $10^{-4}$. The second moment thus controls how strongly a state difference is converted into an update. The first moment sets the direction of the next update, while $\sqrt{v}$ sets how far the update travels along it.

A different data stream instead gives update separation near 1.39 at every checkpoint, more than an order of magnitude above any state perturbation. A state perturbation enters once and then propagates through unchanged maps, whereas a different batch stream changes the map at every step, so the comparison separates divergence seeded by the checkpoint from divergence sustained by continuing inputs. At step 1000 the data-stream condition still shows 8.7\% top-1 disagreement, against at most 1.9\% for the state conditions, and its parameter separation of $3.37\times10^{-3}$ exceeds the state range of $1.1$--$7.1\times10^{-4}$. The divergence therefore persists at the token level as well as in the weights. Appendix~\ref{app:muon} gives the full conditions.

\section{Training Control}
If sensitivity changes across training, stabilization timing should matter. We vary controller strength to measure how much separation can be reduced and at what validation-loss cost, then hold the scheduled activation budget fixed to test whether sensitivity-chosen windows beat gradient- and update-norm schedules. The sweep spans basic, tuned, and strong variants, so it traces a frontier rather than one operating point.

\begin{figure*}[!t]
\centering
\includegraphics[width=\textwidth]{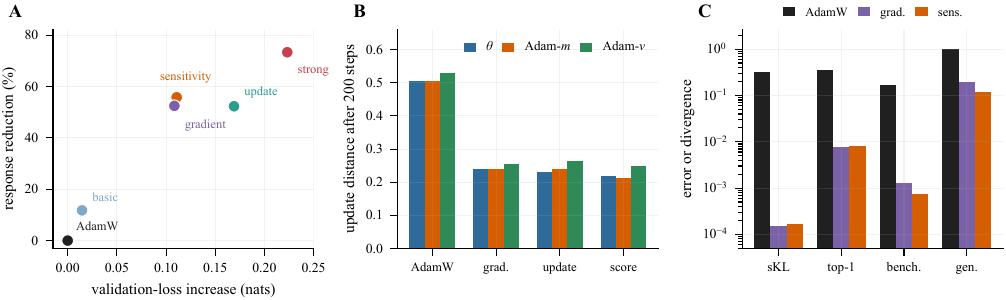}
\caption{Timing of stabilization. A: response reduction versus validation-loss increase. B: matched-budget update separation by state channel. C: output disagreement and generation divergence; lower is better.}
\label{fig:control}
\end{figure*}

\paragraph{Intervention design.}
We use the baseline diagnostic scores to decide when to apply AdamW stabilization. The schedule is offline and replayed in a subsequent training run. The basic variant reduces the effective learning rate and damps first-moment memory; stronger variants also slow second-moment adaptation, damp the preconditioner, and cap update size relative to weight norm.

The design follows the state differential in Equation~\ref{eq:tangent}. Write the update perturbation as $q_{u,t+1}=\sum_{z\in\{\theta,m,v\}}D_{z,t}q_{z,t}$, where $D_{z,t}$ is its linear coefficient with respect to state block $z$. The next weight displacement obeys
\begin{equation}
\begin{split}
\|q_{\theta,t+1}\|\le{}& |1-\eta_t\lambda_w|\|q_{\theta,t}\|\\
&+\eta_t\sum_{z\in\{\theta,m,v\}}\|D_{z,t}\|\|q_{z,t}\|.
\end{split}
\label{eq:main_control_bound}
\end{equation}
The bound separates an inherited weight difference from differences introduced through the next update: reducing the learning rate weakens the latter, while first-moment damping, preconditioner damping, slower second-moment adaptation, and update caps each limit another contribution. Appendix~\ref{app:control} gives the full local-bound argument and all constants.

\paragraph{Matched budgets.}
To test window selection, we keep the intervention strengths and the number of intervals at each strength unchanged, so all schedules use the same activation budget. Sensitivity intervals follow the offline scores; the gradient and update schedules take the checkpoints where that norm is largest, so only the checkpoints at which they act differ. Their scheduled mean activation is 0.0759 and mean learning-rate scale is 0.976. Table~\ref{tab:control} averages 200-step update separation over ten checkpoints and three state channels; its final three rows share this scheduled budget.
\begin{table}[t]
\caption{Controller comparison. Response is 200-step update separation averaged over ten checkpoints and three channels; the last three rows share a scheduled budget.}
\label{tab:control}\centering
\small
\begin{tabular}{lrr}\toprule
Method / schedule & Response & Val. loss\\\midrule
AdamW & 0.514 & 3.919\\
Sensitivity / basic & 0.453 & 3.934\\
Sensitivity / tuned & 0.227 & 4.030\\
Sensitivity / strong & 0.137 & 4.143\\\midrule
Gradient / matched & 0.244 & 4.028\\
Update / matched & 0.245 & 4.088\\
Sensitivity / matched & 0.227 & 4.030\\\bottomrule
\end{tabular}
\end{table}
Separation alone cannot show whether timing helps: suppressing optimization also suppresses separation. Validation loss therefore measures whether the reduction preserves learning progress. Stronger intervention lowers separation to about 0.14 but raises validation loss to 4.143, as Figure~\ref{fig:control}A shows. At matched scheduled strength, the sensitivity schedule instead reduces separation by 7.0\% relative to gradient selection and 7.3\% relative to update selection, with loss between the two, as Table~\ref{tab:control} reports.

\paragraph{Output agreement.}
Table~\ref{tab:control_behavior_main} evaluates a $10^{-5}$ weight perturbation after 200 continuation steps. AdamW, cautious AdamW, and SPAM use their own high-score checkpoints (steps 2000, 2000, and 1750); both matched schedules use 2000. Token metrics average 40 held-out batches; generation uses 128 prompts and 64 greedy tokens; benchmark disagreement covers the full validation sets.

\begin{table}[!t]
\caption{Output agreement after 200 continuation steps at the shared sensitive checkpoint. sKL: symmetric KL; dis.: disagreement; Gen. EM: greedy-generation exact match.}
\label{tab:control_behavior_main}\centering
\footnotesize
\setlength{\tabcolsep}{1.5pt}
\begin{tabular}{lrrrr}\toprule
Method & sKL & Top-1 dis. & Gen. EM & Bench. dis.\\\midrule
AdamW & 0.319993 & 35.71\% & 0.8\% & 16.21\%\\
Cautious AdamW & 0.372854 & 38.55\% & 0.0\% & 18.45\%\\
SPAM & 0.307452 & 34.89\% & 0.0\% & 15.31\%\\\midrule
Gradient schedule & 0.000153 & 0.76\% & 80.5\% & 0.13\%\\
Sensitivity schedule & 0.000167 & 0.83\% & 88.3\% & 0.07\%\\\bottomrule
\end{tabular}
\end{table}

At the shared checkpoint, sensitivity timing reduces top-1 disagreement from 35.7\% to 0.83\% and benchmark disagreement from 16.2\% to 0.07\%, and raises generation exact match from 0.8\% to 88.3\% (Table~\ref{tab:control_behavior_main}, Figure~\ref{fig:control}C). The gradient schedule is close on disagreement but lower on exact match; KL averages steps, whereas exact match scores whole sequences. Appendix comparisons with SPAM \citep{huang2025spam} and cautious AdamW \citep{liang2026cautious} leave top-1 disagreement near 35--39\% while controlling gradient scale, so the properties are distinct.

The two readings also differ in what they average. The 7\% figure spans ten checkpoints and three channels, whereas behavior is measured at the single shared checkpoint. There, small changes settle near-ties and flips concentrate at low-margin positions, so a modest cut in mean separation removes most of the disagreement. The gradient schedule matches this on top-1 (0.76\% versus 0.83\%) but trails by 7.8 points of exact match, damping the wrong window. Strength alone reaches 0.137 but costs 4.143 validation loss, 0.11 nats above the matched schedule.

\section{Conclusion}
We study how small perturbations propagate through pre-training by treating weights and optimizer memory as an evolving dynamical state. A normalized score measures local sensitivity, and paired continuations trace perturbations to logits, generations, and benchmark decisions. At a sensitive checkpoint, a $10^{-5}$ weight perturbation leaves validation loss close yet changes 35.7\% of top-1 predictions and 16.2\% of benchmark answers after 200 steps. Moments matter as much as weights, the pattern transfers to 1B parameters, and offline-timed stabilization reduces top-1 disagreement to 0.83\% while recovering 88.3\% exact generation match at a validation-loss cost of 0.11 nats. The effect persists from 200 to 2000 steps and across amplitudes, so the sensitive windows are not tied to one horizon. Across seeds, scales, and channels, sensitivity concentrates in identifiable windows, is predicted by the score, and can be controlled at small cost.

\clearpage
\section*{Impact Statement}
Modeling language-model pre-training as a dynamical system provides a way to study the effects of small perturbations with tools developed over decades in dynamical systems and control. Our work connects this perspective to optimizer state, measurable sensitivity, and subsequent model behavior. It supports further investigation of stability, bifurcations, and feedback control during pre-training, offering concrete questions about how training regimes change and how interventions shape their trajectories. The broader scientific opportunity is to bring these established analytical methods into the study and design of large-model pre-training.

\FloatBarrier
\bibliography{lyapunov_llm}
\bibliographystyle{icml2026}
\clearpage
\appendix
\setcounter{dbltopnumber}{3}
\renewcommand{\dbltopfraction}{0.85}
\renewcommand{\dblfloatpagefraction}{0.72}

The appendix documents the probe, its validation, and the full result summaries. Appendices~\ref{app:implementation} and~\ref{app:scope} give the implementation and scope; Appendices~\ref{app:predictive}--\ref{app:crossseed_validation} test endpoint prediction; Appendix~\ref{app:local_response} relates local response to output changes; Appendices~\ref{app:derivation}--\ref{app:channels} derive the state map, the relative geometry, and the perturbation channels; and Appendices~\ref{app:settings}--\ref{app:muon} record settings, controller details, full summaries, and the Muon pilot.

\section{Probe Implementation and Interpretation}
\label{app:implementation}

Probe cost is material. On one RTX 3090, a small-model AdamW step takes $71.54\pm0.55$ ms; a three-iteration probe takes $1.002\pm0.007$ s and 11.61 GB, versus 5.80 GB for training. Probing every 500 steps adds about 2.8\% amortized time. Controller steps take $86.56\pm0.10$ ms, a separate 21.0\% overhead. An end-to-end compute comparison must include both costs.

\begin{algorithm}[t]
\caption{Normalized Adam-state checkpoint diagnostic}
\label{alg:probe}
\begin{algorithmic}[1]
\REQUIRE Checkpoint $(\theta,m,v)$, diagnostic batch $B$, window $K$, optimizer settings
\STATE Compute $g=\nabla\ell(\theta;B)$ and retain Hessian--vector products
\STATE Draw $q_\theta\sim\mathcal N(0,I)$, set $q_m=q_v=0$, and normalize the raw state tangent
\FOR{$j=0,\ldots,K-1$}
\STATE Update local moments with Equations~\eqref{eq:adam-m}--\eqref{eq:adam-v}, and their tangents with Equations~\eqref{eq:q-m}--\eqref{eq:q-theta}
\STATE Compute update and log-preconditioner tangents
\STATE Evaluate $n_j=N(\widetilde q_{j+1})+10^{-12}$ using Equation~\ref{eq:norm}
\STATE Record the relative norm and renormalize the next tangent as described in Appendix~\ref{app:implementation}
\ENDFOR
\STATE \textbf{Return} score $S_K$ using Equation~\ref{eq:implemented}
\end{algorithmic}
\end{algorithm}

The diagnostic reports a local tangent response.

The parameter block is initialized with Gaussian noise and the moment blocks with zero. The initial tangent is normalized in the raw extended-state RMS norm, not Equation~\ref{eq:norm}. At local iteration $j$, the code applies the analytical Adam tangent recurrence, evaluates the relative norm $n_j=N(\widetilde q_{j+1})+10^{-12}$, and records it. The reported score uses the last recorded factor,
\begin{equation}
S_K=\frac{\log n_{K-1}}{K}.
\label{eq:implemented}
\end{equation}
Each preceding iteration renormalizes the tangent before the next step, so $n_{K-1}$ measures the relative response of a locally unit-sized tangent over the final Adam step, not the cumulative growth of an unnormalized tangent across the window. The $1/K$ factor gives the score the same per-step scaling as Equation~\ref{eq:ftle}, but $S_K$ is not an estimate of that finite-time exponent: Equation~\ref{eq:ftle} accumulates growth over the entire window, whereas the numerator of $S_K$ is the final step's local response, measured after every preceding iteration has renormalized the tangent. We therefore treat the score as a local ranking statistic for checkpoint comparison and intervention timing, not as a contraction certificate, and validate its association with finite responses empirically.
For the small-model implementation, the tangent passed to the next iteration is $\widetilde q_{j+1}/n_j$. For the model-parallel OLMo implementation, it is normalized by its raw RMS norm instead. OLMo uses $K=1$, so intermediate normalization does not affect its reported single-iteration result. The update-space score is defined analogously; the raw-state variant has its own normalization convention and should not be interpreted as an otherwise identical change of metric.

The small-model probe holds checkpoint parameters, the batch gradient, Hessian action, and learning rate fixed while advancing analytical moment recurrences and bias-correction counters. Parameter denominators use the checkpoint tensors, and update denominators use each local iteration's computed direction. This avoids simulating the nonlinear model trajectory but does not produce the Jacobian product along subsequent realized training states and batches.

Square roots use a $10^{-30}$ second-moment floor. Clipping and optimizer masks are piecewise differentiable, while floating-point rounding and 8-bit moment storage are not represented by the smooth tangent recurrence. OLMo moment buffers are dequantized into FP32 coordinates for the probe. The result describes an analytical local surrogate, not an exact derivative of quantized hardware execution. Finite perturbations must be measured after casting, since sufficiently small changes may be rounded away.

\section{Scope and Further Validation}
\label{app:scope}
The dense predictive comparison uses ten checkpoints from one trajectory. The reported MAE interval includes zero improvement, and the sign-flip test does not establish a statistically reliable advantage. Nearby checkpoints share history. High/low checkpoints also differ in training stage, learning rate, and optimizer state. Replications cover three small-model seeds and one OLMo continuation; repeated amplitudes, channels, and horizons are dependent conditions.

The additional validation uses the existing three training seeds and checkpoints to measure independent perturbation responses and evaluate prediction on a held-out training seed. Repeated directions refine each checkpoint response. SPAM and cautious AdamW test Adam-family variants; a different optimizer family requires its own state and update derivative. Offline schedule comparisons match activation strength, and probe cost is reported separately from training cost.

\section{Endpoint Prediction from Checkpoint Statistics}
\label{app:predictive}
The checkpoint contrasts raise a second explanation: sensitivity may largely follow the size of the training step. We compare the sensitivity score against learning rate, update norm, gradient norm, and curvature using responses from the separate paired continuations; Table~\ref{tab:diagnostic_value} reports the correlations. After averaging continuous-perturbation responses within each of ten checkpoints, learning rate and update norm correlate with final update separation at 0.984 and 0.979, compared with 0.640 for the sensitivity score. For empirical log-update growth, the correlations are 0.872, 0.892, and 0.731. Step-scale statistics thus explain much of the endpoint response.

\begin{figure*}[t]
\centering
\includegraphics[width=\textwidth]{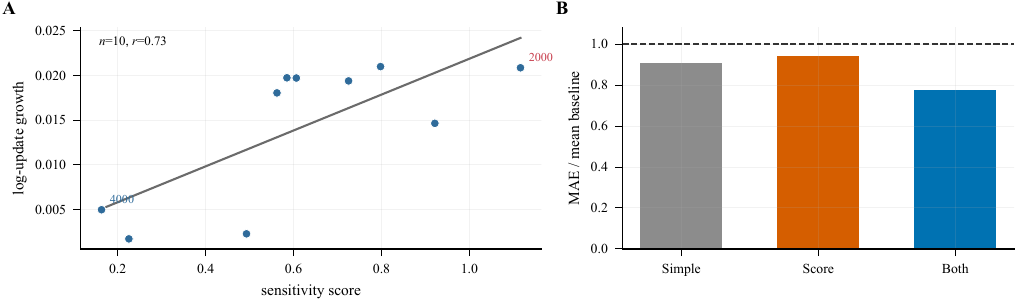}
\caption{Checkpoint statistics and endpoint prediction. A: sensitivity score versus log-update growth. B: leave-one-checkpoint-out MAE relative to the mean baseline; lower is better.}
\label{fig:diagnostic_appendix}
\end{figure*}

The score retains an association with growth after controlling for learning rate and update norm: the partial correlation is 0.692 ($p=0.057$, six degrees of freedom). In nested leave-one-checkpoint-out ridge regression, adding the score to the simple features reduces growth MAE by 14.5\%, as Figure~\ref{fig:diagnostic_appendix}B shows; the bootstrap 95\% interval is $-3.3$\% to 30.7\%, and the paired sign-flip test gives $p=0.143$. The original result concerns growth on this trajectory. Independent cross-seed validation gives target-dependent changes without a consistent advantage over combined simple baselines; Appendix~\ref{app:crossseed_validation} reports the comparison.

\begin{table}[t]
\caption{Pearson correlations of checkpoint statistics with future update separation and growth over ten checkpoints.}
\label{tab:diagnostic_value}
\begin{center}
\footnotesize
\resizebox{\linewidth}{!}{%
\begin{tabular}{lrrr}
\toprule
Diagnostic & Endpoint & Growth & Early growth \\
\midrule
Norm. score & 0.640 & 0.731 & 0.628 \\
Gradient norm & -0.229 & -0.177 & -0.370 \\
Update norm & 0.979 & \textbf{0.892} & 0.979 \\
Learning rate & \textbf{0.984} & 0.872 & \textbf{0.991} \\
Curvature proxy & 0.216 & 0.182 & 0.133 \\
\bottomrule
\end{tabular}}
\end{center}
\end{table}

\section{Independent Cross-Seed Prediction Validation}
\label{app:crossseed_validation}
The additional validation reuses the three existing training trajectories and their ten checkpoints. Three independent parameter-perturbation directions at each checkpoint give 90 paired continuations and 30 checkpoint-level observations. Each continuation lasts 200 steps. The response batches and directions are separate from the diagnostic probe. Both branches use an effective response batch of 16 sequences, accumulated from microbatches of two to bound memory. This numerical cohort is analyzed separately from the original 364 paired trajectories.

Directions are aggregated by the checkpoint median. An outer fold holds out one complete training seed. Standardization and ridge regularization use only the remaining seeds, with inner validation also split by training seed. The combined simple baseline includes gradient norm, update norm, update-to-weight ratio, learning rate, current training loss, and the change in mean training loss between the two preceding 100-step windows. These features are available before the continuation. The augmented predictor adds the existing checkpoint sensitivity score.

\begin{table*}[t]
\caption{Cross-seed incremental prediction. MAE reduction from adding the sensitivity score to the simple feature set; intervals resample whole-seed errors and the last column gives the exact sign-flip test.}
\label{tab:crossseed_validation}
\centering
\begin{tabular}{lrrr}\toprule
Prediction target & MAE reduction & Seed-block interval (\%) & Sign-flip $p$\\\midrule
Early log-update growth & -4.1\% & $[-11.9, -0.6]$ & 0.25 \\
Parameter separation & -3.4\% & $[-7.1, 3.5]$ & 0.75 \\
Update separation & 4.0\% & $[-0.8, 7.8]$ & 0.50 \\
Token disagreement & 6.3\% & $[-0.9, 14.4]$ & 0.50 \\
\bottomrule\end{tabular}
\end{table*}

The interval in Table~\ref{tab:crossseed_validation} resamples entire seed error blocks conditional on fitted out-of-fold predictions. With three trajectories it is descriptive; the last column gives the exact whole-seed sign-flip test. The incremental changes depend on the prediction target. Adding the score reduces the point-estimate errors for final update separation and token disagreement, while increasing errors for early growth and final parameter separation. These three-seed results do not establish a consistent predictive advantage over the combined simple baseline. They complement the original ten-checkpoint analysis, which uses a different growth horizon and feature set.

For the training-stage comparison, checkpoints are matched at exactly the same global step with at most a 1\% relative learning-rate difference. A separate performance-matched analysis adds a 0.05-nat pre-continuation validation-loss caliper. Within each step, the eligible pair with the largest score gap is selected without using subsequent responses. Ten stage pairs satisfy the performance caliper. Their median high-minus-low update-separation difference is 0.00315, whereas the corresponding median token-disagreement difference is $-0.00244$. Thus matched trajectory and behavioral effects must be reported separately. These pairs reuse the same three training trajectories and are descriptive comparisons, not ten independent training replications.

The primary growth target is the log-update-distance slope over steps 1--20. Final parameter separation uses the predefined reference-endpoint normalization $\|\theta_H^{\rm pert}-\theta_H^{\rm ref}\|_2/(\|\theta_H^{\rm ref}\|_2+10^{-12})$. Token behavior is evaluated on four shared validation batches, separately from the original full benchmark and 128-prompt evaluations. The analysis design was refined after partial collection and is not a preregistered experiment.

\section{Local Response and Output Changes}
\label{app:local_response}
For a twice continuously differentiable realized-window map $F$, assume $\|D^2F(z)\|\le L$ on the line segment between $x$ and $x+\delta$. Taylor's theorem gives
\begin{equation}
\|F(x+\delta)-F(x)-DF(x)\delta\|\le \tfrac12L\|\delta\|^2.
\end{equation}
This bounds the error of the tangent response in a fixed norm. It does not justify extrapolating the frozen checkpoint score to a 200-step saturated response.

For held-out loss, $\Delta L=\nabla L^\top\Delta\theta+O(\|\Delta\theta\|^2)$. Logits instead change as $\Delta z=J_z\Delta\theta+O(\|\Delta\theta\|^2)$. A displacement can have a small projection onto the average-loss gradient and still change individual logits. If the original top-logit margin is $\mu$, the top prediction is preserved whenever $\|\Delta z\|_\infty<\mu/2$. Small-margin tokens can therefore flip with small loss changes. Loss and prediction agreement measure different projections of the model change.

\section{Mathematical Construction of Optimizer-Aware Finite-Time Stability}
\label{app:derivation}

This appendix derives the AdamW state map and the geometry underlying the local sensitivity analysis. Adam moments belong to the state, and local response is conditioned on the realized batches and optimizer settings. The relative coordinates specify the distance used to compare displacements.

\subsection{Pre-training as a random non-autonomous dynamical system}

Let $(\Omega,\mathcal F,\mathrm P)$ describe all exogenous randomness in
pre-training.  A realization $\omega\in\Omega$ determines the batch stream,
dropout masks, numerical kernels, collective-reduction order, and finite
precision effects.  Write
\begin{equation}
    \xi_t(\omega)=(B_t,\zeta_t)
\end{equation}
for the batch and remaining numerical randomness at step $t$.  Deterministic
controls such as the learning-rate schedule and optimizer hyperparameters are
collected in
\begin{equation}
    \alpha_t=(\eta_t,\beta_{1,t},\beta_{2,t},\epsilon_t,\lambda_{w,t}).
\end{equation}
For parameters $\theta_t\in\mathbb{R}^d$ and Adam memory
$m_t,v_t\in\mathbb{R}^d$, the endogenous state is
\begin{equation}
    x_t=(\theta_t,m_t,v_t)\in\mathcal X=\mathbb{R}^{3d}.
\end{equation}
The realized training process is the random cocycle
\begin{equation}
    x_{t+1}=T_{t,\omega}(x_t)
    :=T(x_t;\alpha_t,\xi_t(\omega)).
    \label{eq:random-cocycle}
\end{equation}
It is non-autonomous because $\alpha_t$ and $\xi_t$ change with $t$, and stochastic
before conditioning on $\omega$.  Once a realized environment is fixed,
\eqref{eq:random-cocycle} is a deterministic finite sequence of maps.
This conditional object is what our same-batch paired trajectories probe.

Two notions of sensitivity must be separated.  An endogenous state
perturbation changes $x_t$ while holding the future drive fixed:
\begin{equation}
    x_t' = x_t+\delta x_t,\qquad
    \xi_{t:t+K-1}'=\xi_{t:t+K-1}.
    \label{eq:endogenous-coupling}
\end{equation}
An exogenous perturbation, such as batch-order change, instead uses
$x_t'=x_t$ but $\xi_j'\neq\xi_j$ for some $j\ge t$.  Only the first is governed
by the homogeneous tangent map.  The second obeys a forced tangent equation,
derived below.

\subsection{The exact AdamW state map}

For the realized mini-batch loss
$\ell_t(\theta)=\ell(\theta;B_t,\zeta_t)$, define
\begin{equation}
    g_t=\nabla\ell_t(\theta_t),\qquad
    H_t=\nabla^2\ell_t(\theta_t).
\end{equation}
With elementwise product $\odot$, AdamW performs
\begin{align}
    m_{t+1} &= \beta_1m_t+(1-\beta_1)g_t, \\
    v_{t+1} &= \beta_2v_t+(1-\beta_2)(g_t\odot g_t),
    \label{eq:adam-moments-app}\\
    \widehat m_{t+1} &= c_{1,t+1}m_{t+1},\qquad
    c_{1,t+1}=(1-\beta_1^{t+1})^{-1},\\
    \widehat v_{t+1} &= c_{2,t+1}v_{t+1},\qquad
    c_{2,t+1}=(1-\beta_2^{t+1})^{-1},\\
    u_{t+1} &= \widehat m_{t+1}\oslash
    (\sqrt{\widehat v_{t+1}}+\epsilon),
    \label{eq:preconditioned-update-app}\\
    \theta_{t+1} &= (1-\eta_t\lambda_w)\theta_t-\eta_tu_{t+1}.
    \label{eq:adamw-theta-app}
\end{align}
Here $u_{t+1}$ is the preconditioned Adam direction before multiplication by
$\eta_t$.  Decoupled weight decay appears only in
\eqref{eq:adamw-theta-app}; this distinction matters in the Jacobian.

\subsection{Linearization of the extended AdamW state}

Let
\begin{equation}
    q_t=(q_{\theta,t},q_{m,t},q_{v,t})
    =\frac{\partial x_t(r)}{\partial r}|_{r=0}
\end{equation}
be the tangent generated by a differentiable perturbation family
$x_t(r)=x_t+r q_t+o(r)$.  Because
$D_\theta g_t[q_{\theta,t}]=H_tq_{\theta,t}$, define
\begin{equation}
    h_t:=H_tq_{\theta,t}.
    \label{eq:hvp-app}
\end{equation}
The implementation evaluates this Hessian--vector product without constructing
$H_t$. Differentiating the two moment recurrences in
\eqref{eq:adam-moments-app} gives
\begin{align}
    q_{m,t+1}
    &=\beta_1q_{m,t}+(1-\beta_1)h_t,
    \label{eq:tangent-m}\\
    q_{v,t+1}
    &=\beta_2q_{v,t}
      +2(1-\beta_2)\operatorname{Diag}(g_t)h_t.
    \label{eq:tangent-v}
\end{align}
Bias-correction factors are state-independent, hence
\begin{equation}
    q_{\widehat m,t+1}=c_{1,t+1}q_{m,t+1},\qquad
    q_{\widehat v,t+1}=c_{2,t+1}q_{v,t+1}.
\end{equation}

To differentiate the adaptive direction, let
\begin{equation}
    r_{t+1}=(\sqrt{\widehat v_{t+1}}+\epsilon)^{-1},
    \qquad u_{t+1}=\widehat m_{t+1}\odot r_{t+1}.
\end{equation}
Coordinatewise differentiation yields
\begin{align}
    q_{r,t+1}
    &=-\frac{q_{\widehat v,t+1}}
    {2\sqrt{\widehat v_{t+1}}
    (\sqrt{\widehat v_{t+1}}+\epsilon)^2},\\
    q_{u,t+1}
    &=r_{t+1}\odot q_{\widehat m,t+1}
    -\frac{\widehat m_{t+1}\odot q_{\widehat v,t+1}}
    {2\sqrt{\widehat v_{t+1}}
    (\sqrt{\widehat v_{t+1}}+\epsilon)^2}.
    \label{eq:tangent-update}
\end{align}
All square roots use a small positive floor in computation.  Finally,
\begin{equation}
    q_{\theta,t+1}
    =(1-\eta_t\lambda_w)q_{\theta,t}-\eta_tq_{u,t+1}.
    \label{eq:tangent-theta}
\end{equation}
Equations~\eqref{eq:tangent-m}--\eqref{eq:tangent-theta} are the recurrence
implemented by the single-device and model-parallel probes.

For clarity, let $R_t=\operatorname{Diag}(r_{t+1})$ and
\begin{equation}
    S_t=\operatorname{Diag}\!(
    \frac{\widehat m_{t+1}}
    {2\sqrt{\widehat v_{t+1}}
    (\sqrt{\widehat v_{t+1}}+\epsilon)^2}).
\end{equation}
Then the update-direction differential is
\begin{equation}
\begin{split}
 q_{u,t+1}={}&R_tc_{1,t+1}
 [\beta_1q_{m,t}+(1-\beta_1)H_tq_{\theta,t}]\\
 &-S_tc_{2,t+1}[\beta_2q_{v,t}\\
 &\qquad+2(1-\beta_2)\operatorname{Diag}(g_t)H_tq_{\theta,t}].
\end{split}
 \label{eq:qu-expanded}
\end{equation}
This expression exposes three amplification mechanisms: curvature $H_t$,
first-moment memory, and the second-moment/preconditioner factor $S_t$.
When coordinates of $\widehat v$ are small, $S_t$ can be large even if the loss
curve is smooth.  This is why parameter-only stability is insufficient for
AdamW.

\subsection{The block Jacobian and finite-time propagator}

Equations above define a block Jacobian $J_t=D_xT_{t,\omega}(x_t)$ such that
$q_{t+1}=J_tq_t$.  Writing $D_{\theta,t},D_{m,t},D_{v,t}$ for the three linear
coefficients in \eqref{eq:qu-expanded},
\begin{equation}
    q_{u,t+1}=D_{\theta,t}q_{\theta,t}
    +D_{m,t}q_{m,t}+D_{v,t}q_{v,t},
\end{equation}
the block Jacobian $J_t$ is specified without materializing a matrix by
\begin{align}
[J_tq]_{\theta}
&=[(1-\eta_t\lambda_w)I-\eta_tD_{\theta,t}]q_\theta\\
&\quad-\eta_tD_{m,t}q_m-\eta_tD_{v,t}q_v,\\
[J_tq]_{m}&=(1-\beta_1)H_tq_\theta+\beta_1q_m,\\
[J_tq]_{v}&=2(1-\beta_2)\operatorname{Diag}(g_t)H_tq_\theta+\beta_2q_v.
\label{eq:block-jacobian}
\end{align}
The $K$-step tangent propagator conditional on the realized drive is
\begin{equation}
    \Phi_{t,K}(\omega)
    =J_{t+K-1}(\omega)\cdots J_t(\omega),
    \qquad q_{t+K}=\Phi_{t,K}q_t.
    \label{eq:tangent-propagator}
\end{equation}
No stationarity or asymptotic limit is required.

For a discrete exogenous change, define its finite one-step injection along the
reference trajectory by
\begin{equation}
    d_t^{\mathrm{exog}}:=T(x_t;\alpha_t,\xi_t')-
    T(x_t;\alpha_t,\xi_t).
\end{equation}
Linearizing only with respect to the continuous state displacement gives the
first-order forced response
\begin{equation}
\Delta x_{t+K}\simeq\Phi_{t,K}\Delta x_t+
\sum_{j=0}^{K-1}\Phi_{t+j+1,K-j-1}d_{t+j}^{\mathrm{exog}}.
\label{eq:forced-response}
\end{equation}
Thus batch-order experiments measure both injection magnitude and subsequent
amplification.  They are displayed separately from homogeneous state perturbations in
our figures.

\section{Optimizer-Relative Geometry and Finite-Time Growth}
\label{app:geometry}

The raw extended state adds parameters, moments, and squared moments, so its Euclidean norm changes with loss scaling and unit choice. The following sections define the layer-balanced coordinates, the resulting optimizer-relative norm, and the directional exponents computed in it, then connect tangent growth to finite perturbations.

\subsection{Coordinate dependence of raw Euclidean state distance}

The raw norm
$\|q\|_2^2=\|q_\theta\|_2^2+\|q_m\|_2^2+\|q_v\|_2^2$
adds parameters, gradients, and squared gradients and changes with the choice of
units. If the loss is rescaled by $c>0$, then approximately
$m\mapsto cm$, $v\mapsto c^2v$, while the Adam direction
$\widehat m/(\sqrt{\widehat v}+\epsilon)$ is nearly unchanged away from the
$\epsilon$ floor.  A raw state norm can therefore change by orders of magnitude
without a comparable change in optimizer behavior.  The negative raw-state
score in our ablation reflects this coordinate dependence alongside update-space
expansion.

\subsection{Layer-balanced relative coordinates}

Partition parameters into tensors or layers $\mathcal L$.  For aligned
collections $a=(a_\ell)_{\ell\in\mathcal L}$ and
$b=(b_\ell)_{\ell\in\mathcal L}$, define
\begin{equation}
    \mathcal R(a;b)^2
    =\frac{1}{|\mathcal L|}\sum_{\ell\in\mathcal L}
    \frac{\operatorname{RMS}(a_\ell)^2}
    {\operatorname{RMS}(b_\ell)^2+\varepsilon_{\mathrm{floor}}^2},
    \label{eq:layer-relative}
\end{equation}
where $\varepsilon_{\mathrm{floor}}$ is a numerical floor. Equal averaging over tensors prevents a
large embedding matrix from automatically dominating every smaller
normalization or projection tensor.

The preconditioner coordinate is
\begin{equation}
    \rho_{t+1}=\log(\sqrt{\widehat v_{t+1}}+\epsilon).
\end{equation}
Its tangent is
\begin{equation}
    q_{\rho,t+1}
    =\frac{q_{\widehat v,t+1}}
    {2\sqrt{\widehat v_{t+1}}
    (\sqrt{\widehat v_{t+1}}+\epsilon)}.
    \label{eq:rho-tangent}
\end{equation}
This is dimensionless and directly measures multiplicative change in the
diagonal coordinate system used by Adam.

Our optimizer-relative tangent norm is
\begin{equation}
\begin{split}
\|q_t\|_{\mathrm{opt}}^2
={}&w_\theta\mathcal R(q_{\theta,t};\theta_t)^2
+w_u\mathcal R(q_{u,t};u_t)^2\\
&+w_\rho\frac{1}{|\mathcal L|}\sum_{\ell\in\mathcal L}
\operatorname{RMS}(q_{\rho,t}^{(\ell)})^2.
\end{split}
\label{eq:optimizer-norm}
\end{equation}
The reported experiments use equal weights.  The three terms answer distinct
questions: relative motion of the model, relative motion of the direction that
will update it, and deformation of Adam's preconditioning geometry.

\subsection{Directional and maximal finite-time exponents}

For a nonzero initial tangent $q_t$, the directional finite-time exponent is
\begin{equation}
    \lambda_{t,K}(q_t;\omega)
    =\frac{1}{K}\log
    \frac{\|\Phi_{t,K}(\omega)q_t\|_{\mathrm{opt},t+K}}
    {\|q_t\|_{\mathrm{opt},t}}.
    \label{eq:directional-ftle}
\end{equation}
The maximal exponent in this metric is
\begin{equation}
    \lambda_{t,K}^{\max}(\omega)
    =\frac{1}{K}\log
    \|M_{t+K}^{1/2}\Phi_{t,K}M_t^{-1/2}\|_2,
    \label{eq:max-ftle}
\end{equation}
where $M_t$ is the local positive metric associated with
\eqref{eq:optimizer-norm}.  Computing
\eqref{eq:max-ftle} exactly is infeasible at LLM scale.  Matrix-free directional multiplication evaluates parameter-initiated displacements through the extended dynamics. Channel-specific nonlinear replays separately initialize
perturbations in $\theta$, $m$, and $v$. At LLM dimension and short $K$, a
random-direction probe characterizes typical parameter-seeded expansion;
maximal expansion requires singular-vector optimization.
Multiple tangent seeds test directional robustness; multiple diagnostic
batches test dependence on the realized local map. Appendix~\ref{app:implementation} defines the score reported in the experiments.

For the conventional realized-window exponent, metric renormalization gives the Benettin-style construction
\begin{align}
    \widetilde q_{j+1}&=J_{t+j}q_j,\\
    r_j&=\|\widetilde q_{j+1}\|_{\mathrm{opt},t+j+1},\\
    q_{j+1}&=\widetilde q_{j+1}/r_j,\\
    \widehat\lambda_{t,K}^{\mathrm{state}}
    &=\frac{1}{K}\sum_{j=0}^{K-1}\log r_j.
    \label{eq:benettin}
\end{align}
The corresponding mathematical update-space projection replaces the full norm by
\begin{equation}
    \|q_t\|_{u}=\mathcal R(q_{u,t};u_t),
    \qquad
    \widehat\lambda_{t,K}^{\mathrm{update}}
    =\frac1K\sum_{j=0}^{K-1}\log r_{u,j}.
    \label{eq:update-ftle-app}
\end{equation}
The raw-state projection uses the RMS norm of $(q_\theta,q_m,q_v)$. The score used in the reported ablations is defined in Appendix~\ref{app:implementation}.

\subsection{Frozen-local and realized-window interpretations}

The ideal realized-window propagator uses the sequence
$J_t,J_{t+1},\ldots,J_{t+K-1}$ evaluated on subsequent batches and states.  The
small-model checkpoint diagnostic instead uses a short frozen-local probe: it
reuses the checkpoint loss geometry while advancing the analytical Adam memory
recurrence. It estimates repeated local expansion of the checkpoint map
conditional on one batch. The OLMo
model-parallel diagnostic uses $K=1$, for which frozen-local and one-step
realized interpretations coincide. Its saved AdamW8bit moment buffers are
dequantized into FP32 tangent coordinates before the analytical recurrence. Our paired 200--2000-step trajectories are
the nonlinear test of whether the local diagnostic transfers beyond this
short probe.

The distinction also explains why the conventional exponent at a checkpoint
is not unique before specifying a diagnostic batch.  Formally,
\begin{equation}
    \lambda_{t,K}=\lambda_{t,K}(q_t;\xi_{t:t+K-1}).
\end{equation}
Averaging across tangent seeds estimates directional uncertainty conditional
on a batch; averaging across diagnostic batches estimates variability across
local realized maps.  Neither repeated perturbation amplitudes nor different
perturbation channels create additional independent checkpoint score samples.

\subsection{From tangent growth to finite perturbation response}

The link between tangent growth and paired trajectories follows from Taylor's theorem.
Assume the $K$-step map
$F_{t,K}=T_{t+K-1}\circ\cdots\circ T_t$ is twice continuously differentiable
in a neighborhood of $x_t$ and
$\sup_z\|D^2F_{t,K}(z)\|\le L_{t,K}$.  Then
\begin{equation}
F_{t,K}(x_t+\delta x)-F_{t,K}(x_t)
=\Phi_{t,K}\delta x+R_{t,K}(\delta x),
\end{equation}
with
\begin{equation}
    \|R_{t,K}(\delta x)\|
    \le \frac{L_{t,K}}{2}\|\delta x\|^2.
    \label{eq:taylor-remainder}
\end{equation}
Consequently, for sufficiently small $\varepsilon=\|\delta x\|$,
\begin{equation}
\frac{\|F_{t,K}(x_t+\delta x)-F_{t,K}(x_t)\|}
{\|\delta x\|}
=\frac{\|\Phi_{t,K}\delta x\|}{\|\delta x\|}+O(\varepsilon).
\label{eq:linear-response}
\end{equation}
Thus the local tangent growth rate predicts the initial log-distance growth. Once the quadratic remainder is non-negligible,
trajectories may cross a threshold, saturate, or enter another basin.  This is
why our score comparison is strongest against empirical log-distance slope,
while the amplitude sweep separately identifies the nonlinear regime.

\section{Perturbation Channels and Observable Projections}
\label{app:channels}

The experiments perturb one state block at a time and compare the result against a batch-order change. This section defines those perturbations and projects the resulting displacement onto loss, logits, and discrete decisions.

\subsection{State perturbations used in the experiments}

For a state component $z\in\{\theta,m,v\}$, a random directional perturbation
is constructed as
\begin{equation}
    z'=z+\varepsilon\|z\|_2\frac{r}{\|r\|_2},
    \qquad r\sim\mathcal N(0,I),
    \label{eq:random-perturbation}
\end{equation}
with the other two components unchanged.  The achieved relative distance is
always measured after the perturbation is represented in the actual numerical
dtype.  For second moments, nonnegativity is enforced after perturbation.  The
three channels correspond to initial tangents
\begin{equation}
q_t^{(\theta)}=(r,0,0),\qquad
q_t^{(m)}=(0,r,0),\qquad
q_t^{(v)}=(0,0,r).
\end{equation}
They excite different columns of the block Jacobian
\eqref{eq:block-jacobian}, but all can reach $q_u$ and subsequently
$q_\theta$.  This formally explains why moment perturbations can be as
consequential as parameter perturbations.

Batch-order perturbation has $q_t=0$ and enters through the forcing term in
\eqref{eq:forced-response}.  Its gradient/update directional distance
can approach $\sqrt{2}$ when two high-dimensional directions decorrelate:
for unit vectors $a,b$ with $a^\top b\approx0$,
\begin{equation}
    \|a-b\|_2=\sqrt{2-2a^\top b}\approx\sqrt2.
\end{equation}

\subsection{Why aggregate loss can remain close}

Let $L(\theta)$ be held-out loss.  For a parameter displacement
$\Delta\theta$,
\begin{equation}
L(\theta+\Delta\theta)-L(\theta)
=\nabla L(\theta)^\top\Delta\theta
+\frac12\Delta\theta^\top\nabla^2L(\bar\theta)\Delta\theta.
\label{eq:loss-projection}
\end{equation}
Even a large trajectory displacement can have small loss projection when it is
nearly orthogonal to the held-out gradient or lies in a low-curvature
subspace.  Loss is therefore a scalar projection of a high-dimensional
difference and provides a scalar projection of the training trajectories.

For an input $x$, let logits be $z_\theta(x)$ and probabilities
$p_\theta(\cdot\mid x)=\operatorname{softmax}(z_\theta(x))$.  Locally,
\begin{equation}
    \Delta z(x)=J_z(x)\Delta\theta+O(\|\Delta\theta\|^2).
\end{equation}
The symmetric KL has the second-order form
\begin{equation}
\operatorname{SKL}(p_\theta,p_{\theta+\Delta\theta})
=\Delta z^\top
[\operatorname{Diag}(p)-pp^\top]\Delta z
+O(\|\Delta z\|^3),
\label{eq:kl-fisher}
\end{equation}
up to the conventional factor determined by the symmetric-KL definition.
This retains many directions that a dataset-averaged scalar loss cancels.

Discrete predictions are controlled by margins.  If $i^*$ is the top logit and
\begin{equation}
    \mu(x)=z_{i^*}(x)-\max_{j\ne i^*}z_j(x),
\end{equation}
then the top-1 prediction is guaranteed unchanged whenever
\begin{equation}
    \|\Delta z(x)\|_\infty<\mu(x)/2.
    \label{eq:margin-condition}
\end{equation}
Examples with small margins can therefore change class under perturbations
that barely move mean loss or aggregate accuracy.  Generation compounds this
effect autoregressively: after the first token disagreement, the two models
condition on different prefixes, adding an exogenous forcing term at every
later decoding step. For a multiple-choice candidate $c$, let
$S_c=\sum_{j=1}^{T_c}\log p(w_j^c\mid x,w_{<j}^c)$ and let
$\mu_{\mathrm{seq}}=S_{c^*}-\max_{c\ne c^*}S_c$. The selected candidate remains
unchanged whenever
$2\max_c\sum_{j=1}^{T_c}|\Delta\log p(w_j^c\mid x,w_{<j}^c)|
<\mu_{\mathrm{seq}}$. Thus token-level probability changes accumulate against a
sequence-score margin. Equations~\eqref{eq:loss-projection}--\eqref{eq:margin-condition}
motivate the experimental observable hierarchy from loss to KL, token churn,
generation divergence, and benchmark answer churn.

\section{Experimental Settings and Existing Result Inventory}
\label{app:settings}

This section records the full training and controller configuration and the coverage of every perturbation cohort. Both learning-rate schedules in Table~\ref{tab:hyperparameters} use linear warmup followed by cosine decay to 0.1 of the peak learning rate. Table~\ref{tab:coverage} lists each cohort with its checkpoints, channels, amplitudes, horizons, and paired continuations; repeated channels and amplitudes are controlled conditions within a training run, not additional independent runs. Table~\ref{tab:control_hyperparameters} gives the controller constants, and both matched schedules use the tuned row.
\begin{table*}[t]
\caption{Training configurations. Effective batch sizes include gradient accumulation.}
\label{tab:hyperparameters}
\begin{center}
\normalsize
\setlength{\tabcolsep}{5pt}
\begin{tabular}{@{}>{\raggedright\arraybackslash}p{0.22\linewidth}
                    >{\raggedright\arraybackslash}p{0.34\linewidth}
                    >{\raggedright\arraybackslash}p{0.37\linewidth}@{}}
\toprule
Configuration & 60M trajectory & OLMo-1B continual pre-training \\
\midrule
Checkpoint / architecture & 8 layers, $d=512$, 8 heads, MLP 1376 & OLMo-2-0425-1B; 16 layers, $d=2048$, 16 heads, MLP 8192 \\
Implemented parameters & 76.8M & 1.485B \\
Data / context & C4 / 256 & Dolma v1.6 sample / 1024 \\
Microbatch $\times$ accumulation & $16\times32$ (131,072 tokens/step) & $1\times32$ (32,768 tokens/step) \\
Steps / warmup & 5000 / 1000 & 2000 / 100 \\
Peak LR / minimum ratio & $8\times10^{-4}$ / 0.1 & $2\times10^{-5}$ / 0.1 \\
Adam $(\beta_1,\beta_2,\epsilon,\lambda_w)$ & $(0.9,0.999,10^{-8},0.1)$ & $(0.9,0.95,10^{-8},0.1)$ \\
Precision / optimizer & BF16 AMP / AdamW & BF16 / bitsandbytes AdamW8bit \\
Gradient clipping & 1.0 & 1.0 \\
\bottomrule
\end{tabular}
\end{center}
\end{table*}

For the expanded control evaluation, generation prompts are a seed-2027
stratified sample of 43 PIQA contexts, 43 HellaSwag contexts, and 42 ARC-Easy
contexts with answer choices and labels removed. Greedy decoding produces 64 new
tokens. Multiple-choice evaluation uses length-normalized candidate sequence
log-likelihood on every validation example. We report exact flip counts and
Wilson intervals in the accompanying JSONL files.

The controller score at checkpoint $t$ is held over the following 500-step interval. The sensitivity schedule is active on intervals beginning at steps 1500, 2000, 2500, and 3500. Matched schedules assign the same effective strengths according to gradient or update norm. After each active optimizer step, moment damping multiplies the first moment by $\max(0.05,1-\gamma_m a_t)$.

\begin{table*}[t]
\caption{Controller settings. Rows increase stabilization strength; columns give the activation threshold and update modifications. Matched schedules use the tuned setting.}
\label{tab:control_hyperparameters}
\begin{center}
\scriptsize
\resizebox{\linewidth}{!}{
\begin{tabular}{lrrrrrrrr}
\toprule
Variant & $\tau$ & $\gamma$ & $c_{\min}$ & $\gamma_m$ & $\gamma_v$ & $\gamma_p$ & $\epsilon$ boost & $c_u$ \\
\midrule
Basic & 0.70 & 1.00 & 0.50 & 0.05 & 0 & 0 & 0 & 0 \\
Internal-tuned / matched & 0.70 & 0.35 & 0.80 & 0.03 & 3.0 & 0.20 & 200 & $6.0\times10^{-4}$ \\
Internal & 0.60 & 0.50 & 0.70 & 0.08 & 8.0 & 0.50 & 1000 & $4.2\times10^{-4}$ \\
\bottomrule
\end{tabular}}
\end{center}
\end{table*}

\begin{table*}[t]
\caption{Experimental coverage. Each row lists a cohort, its perturbation conditions, and its paired continuations.}
\label{tab:coverage}
\begin{center}
\scriptsize
\resizebox{\linewidth}{!}{
\begin{tabular}{lllllll}
\toprule
System / role & Score points & Intervention states & Channels & Amplitudes & Horizons & Pairs \\
\midrule
60M dense map & 10 & 500--5000 & $\theta,m,v$, order & $10^{-6},10^{-5},10^{-4}$ & 200; 500/1000/2000 & 118 \\
60M replications & $2\times10$ & run-specific high/low & $\theta,m,v$ & $10^{-5}$ & 200/1000 & 16 \\
OLMo-1B transfer & 8 & 750/1000/1250/2000 & $\theta,m,v$, order & $10^{-6},10^{-5}$ & 200/1000 & 56 \\
Stability controls & $5\times10$ & 500--5000 & $\theta,m,v$ & $10^{-5}$ & 200 & 150 \\
$0.8\times$-LR AdamW & 10 & run-specific high/low & $\theta,m,v$ & $10^{-5}$ & 200/1000 & 8 \\
SPAM / C-AdamW & $2\times20$ & run-specific high/low & $\theta,m,v$ & $10^{-5}$ & 200/1000 & 16 \\
\bottomrule
\end{tabular}}
\end{center}
\end{table*}

\section{From Diagnosis to Stability Control}
\label{app:control}

The main text uses the checkpoint score as an offline schedule. This section writes the controller, identifies the terms in the state differential that each operation targets, and derives the local bound used to choose them.

Let $\widehat\lambda_t$ be the piecewise-constant score from an offline baseline
run and
define activation
\begin{equation}
    a_t=(\widehat\lambda_t-\tau)_+.
\end{equation}
Our controller modifies factors that appear explicitly in the Jacobian
\eqref{eq:block-jacobian}.  The effective learning-rate scale is
\begin{equation}
    c_t=\max\left(c_{\min},\frac{1}{1+\gamma a_t}\right),
    \qquad \eta_t^{\mathrm{eff}}=c_t\eta_t.
\end{equation}
The first moment is damped by
\begin{equation}
    m_t\leftarrow\max(0.05,1-\gamma_m a_t)m_t,
\end{equation}
and second-moment adaptation is slowed via
\begin{equation}
1-\beta_{2,t}^{\mathrm{eff}}
=\frac{1-\beta_2}{1+\gamma_v a_t}.
\end{equation}
The numerical floor and preconditioner denominator are enlarged in sensitive
windows, reducing the magnitude of $R_t$ and $S_t$ in
\eqref{eq:qu-expanded}. Finally, the preconditioned Adam displacement
$\Delta\theta_t^{\mathrm{Adam}}$ before decoupled weight decay obeys the cap
\begin{equation}
    \|\Delta\theta_t^{\mathrm{Adam}}\|_2
    \le c_u\|\theta_t\|_2.
\end{equation}

These operations target identifiable terms in a local upper bound. For any induced norm,
\begin{equation}
\begin{split}
\|q_{\theta,t+1}\|
\le{}& |1-\eta_t\lambda_w|\|q_{\theta,t}\|\\
&+\eta_t\bigl(\|D_{\theta,t}\|\|q_{\theta,t}\|
+\|D_{m,t}\|\|q_{m,t}\|\\
&\qquad+\|D_{v,t}\|\|q_{v,t}\|\bigr).
\end{split}
\label{eq:control-bound}
\end{equation}
At a fixed local state, learning-rate scaling reduces the update-mediated terms. Moment damping changes persistent $q_m$, slower $v$ adaptation reduces gradient injection into $q_v$, and denominator damping acts on the factors in $D_{m,t}$ and $D_{v,t}$. Their combined effect is measured by the intervention experiments. The
validation-loss cost arises because the same operations can also reduce useful
optimization progress.  Hence the theoretically appropriate evaluation is a
stability--optimization Pareto frontier, exactly the comparison reported in
the control experiment. Table~\ref{tab:control_channels} breaks the response
down by perturbation channel. The ordering of variants is the same in all
three channels, so the controller does not act on one state block alone; the
validation-loss column is the cost side of the frontier.

\begin{table*}[t]
\caption{Controller responses by perturbation channel. Lower separation is better.}
\label{tab:control_channels}
\begin{center}
\small
\resizebox{\linewidth}{!}{
\begin{tabular}{lrrrr}
\toprule
Optimizer & $\theta$ update dist. & Adam-$m$ & Adam-$v$ & Final val loss \\
\midrule
AdamW baseline & 0.505 & 0.507 & 0.529 & 3.919 \\
Sensitivity schedule (basic) & 0.443 & 0.441 & 0.475 & 3.934 \\
Sensitivity schedule (internal-tuned) & 0.219 & 0.213 & 0.249 & 4.030 \\
Sensitivity schedule (internal) & 0.128 & 0.131 & 0.152 & 4.143 \\
Gradient schedule (matched) & 0.239 & 0.239 & 0.254 & 4.028 \\
Update schedule (matched) & 0.231 & 0.240 & 0.264 & 4.088 \\
\bottomrule
\end{tabular}}
\end{center}
\end{table*}
\section{Full Response and Behavior Summaries}
\label{app:additional}

This section collects the shared-input summaries behind Figures~\ref{fig:phenomenon}--\ref{fig:control}. It is organized by claim rather than by run: amplitude and horizon effects, the full response matrix, optimizer and channel comparisons, scale transfer, and replication. Repeated conditions are averaged within a checkpoint, and none of the panels adds an independent training run.

\subsection{Amplitude and long-horizon response}

\begin{figure*}[t]
\begin{center}
\includegraphics[width=\linewidth]{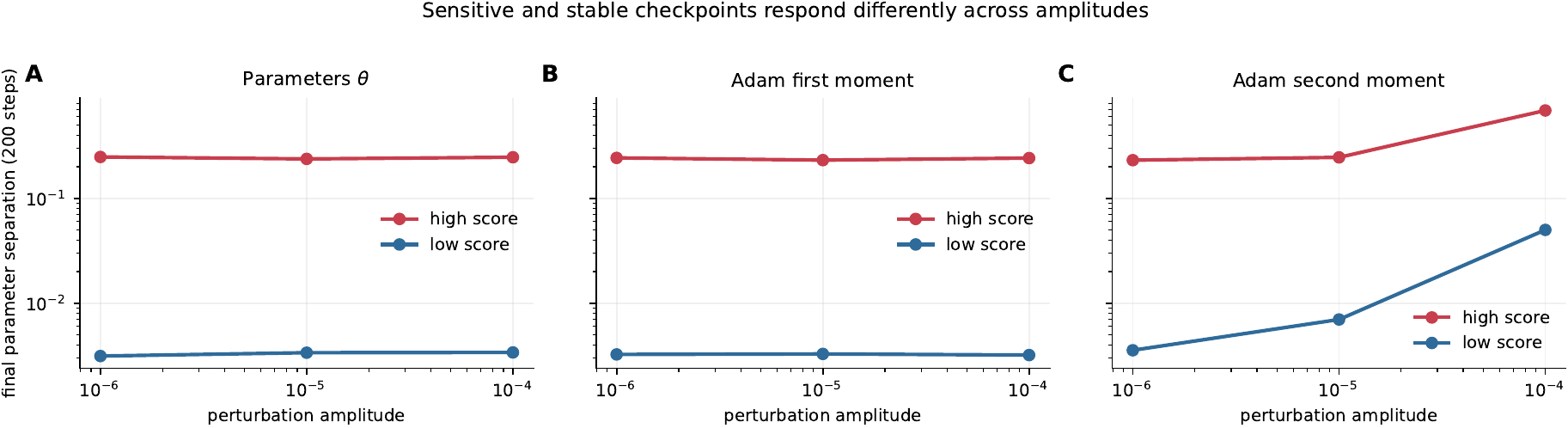}
\end{center}
\caption{Parameter separation at high- and low-score checkpoints. A--C: weight, first-moment, and second-moment perturbations.}
\label{fig:amplitude_main}
\end{figure*}

Figure~\ref{fig:amplitude_main} reports endpoint separation for weight, first-moment, and second-moment perturbations at the selected high- and low-sensitivity checkpoints. Separation orders the two checkpoints at every amplitude. Weight and first-moment parameter separation changes little across the sweep, whereas second-moment parameter separation rises from 0.113 at $10^{-6}$ to 0.818 at $10^{-4}$, while its 200-step update separation stays nearly flat. The flat update response marks saturation, which is why the analysis also measures growth during the continuation.

\begin{figure*}[t]
\begin{center}
\includegraphics[width=\linewidth]{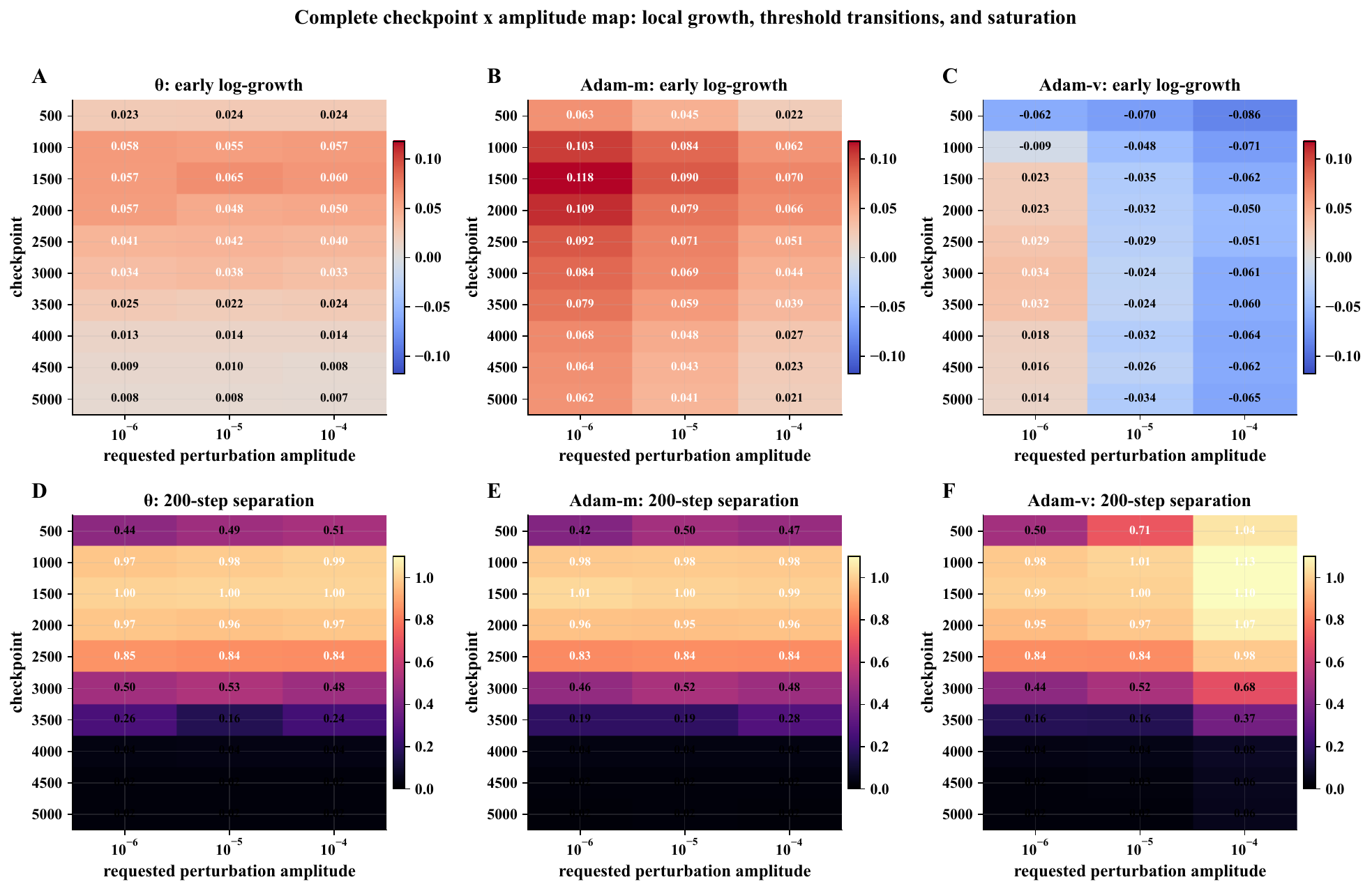}
\end{center}
\caption{Full amplitude response. Color denotes checkpoint score; rows show early growth and endpoint separation.}
\label{fig:amplitude_full}
\end{figure*}

Figure~\ref{fig:amplitude_full} repeats the sweep across all ten checkpoints, with color giving the checkpoint score and rows separating early growth from endpoint separation. Pooled across channels, growth follows the score more consistently than endpoint separation does, and the second-moment column saturates at the largest amplitude. We therefore use growth as the primary target and treat endpoint separation as a downstream readout.

\begin{figure*}[t]
\begin{center}
\includegraphics[width=\linewidth]{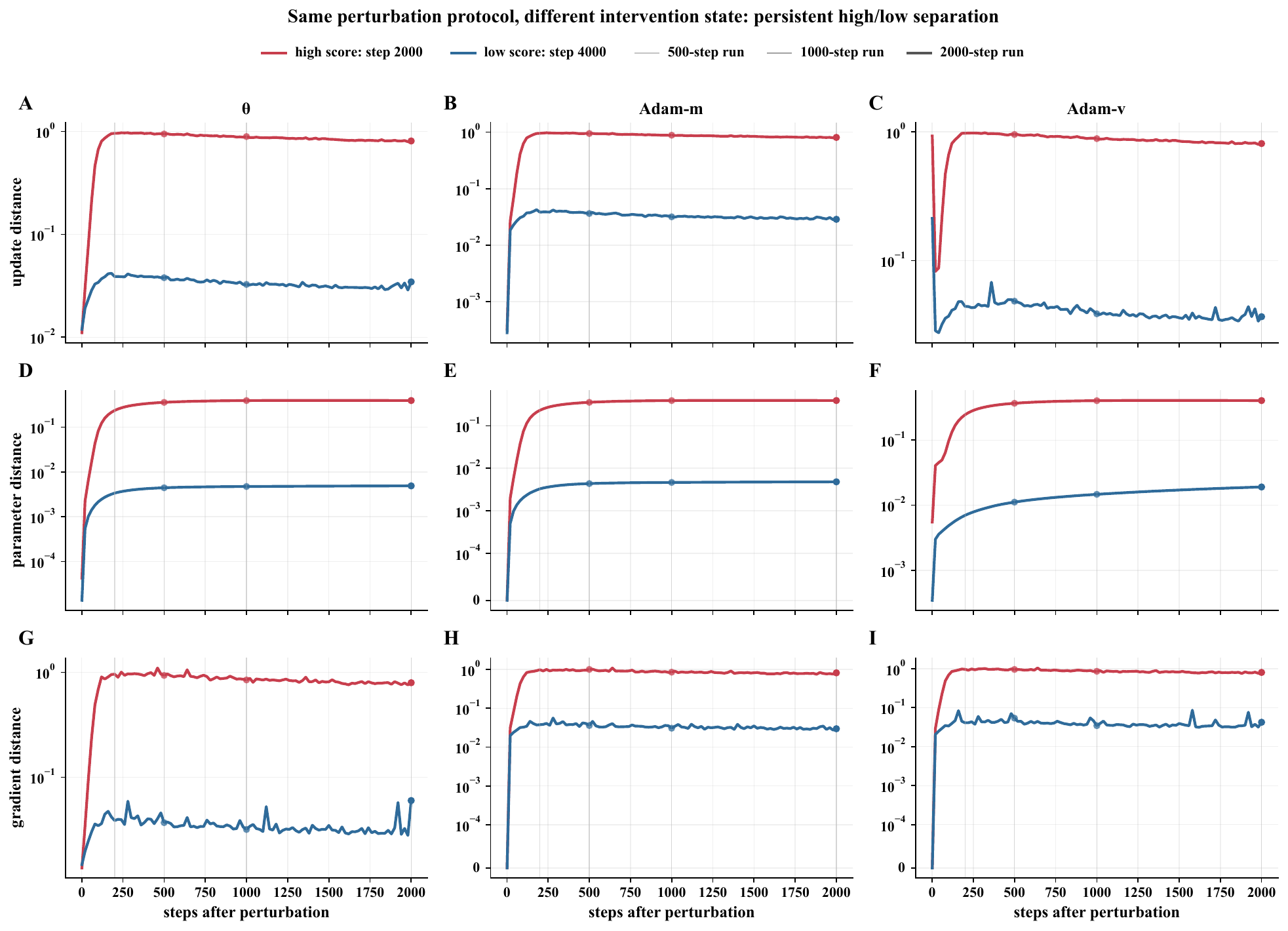}
\end{center}
\caption{Long-horizon responses by state channel for high- and low-score launches.}
\label{fig:long_window}
\end{figure*}

Figure~\ref{fig:long_window} follows the two selected checkpoints over 200, 500, 1000, and 2000 steps. The high-checkpoint curves stay above the low-checkpoint curves in every channel and at every horizon. At 1000 steps, parameter separation reaches 0.39--0.40 from the high checkpoint and stays below 0.015 from the low one. The ordering persists at 2000 steps, so the contrast is not a transient of the 200-step window.

\subsection{Full response matrix}

\begin{figure*}[t]
\begin{center}
\includegraphics[width=\linewidth]{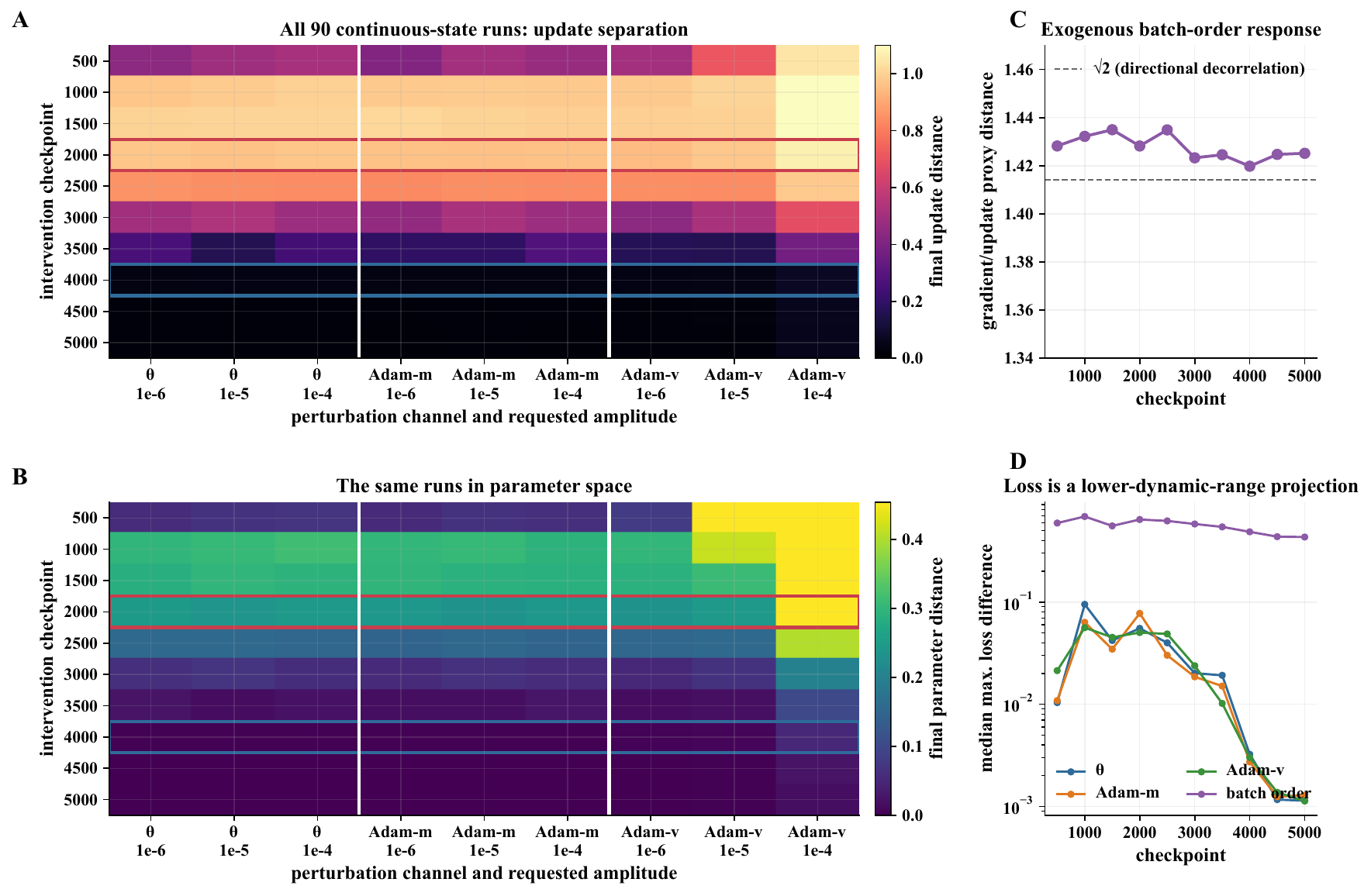}
\end{center}
\caption{Full small-model response matrix. A--B: state-perturbation responses; C: batch-order responses; D: loss differences.}
\label{fig:perturbation_matrix}
\end{figure*}

Figure~\ref{fig:perturbation_matrix} gives the complete small-model response matrix. Panels A--B cover update and parameter separation for the three state channels across ten checkpoints and three amplitudes, panel C gives the batch-order response, and panel D gives held-out loss differences. The state channels produce comparable update separation at $10^{-5}$, which means a checkpoint's sensitivity cannot be read from its weights alone. Batch-order changes instead hold update separation near 1.43 at every checkpoint while state perturbations produce loss differences below 0.06 nats. The contrast separates a persistent input-stream effect from a state perturbation that the training dynamics amplify.

\subsection{Optimizer and channel comparisons}

Table~\ref{tab:optimizer_variants} compares AdamW, cautious AdamW, and SPAM at each optimizer's own selected high- and low-score checkpoints. Gradient norm gives the post-warmup mean and 95th percentile, and response averages final update separation across the three state channels. All three optimizers show a positive high-to-low response ratio, and SPAM reaches the lowest validation loss and gradient norm. The selected checkpoints differ across optimizers, so the table does not isolate the optimizer choice from checkpoint selection.

\begin{table*}[t]
\caption{Optimizer comparison. Gradient norm reports the post-warmup mean and 95th percentile; response compares high- and low-score launches for each optimizer.}
\label{tab:optimizer_variants}
\begin{center}
\small
\setlength{\tabcolsep}{4pt}
\begin{tabular}{lrrrr}
\toprule
Optimizer & Val. loss & Grad. norm & Score mean [range] & Response high / low \\
\midrule
AdamW & 3.919 & 0.287 / 0.373 & 0.620 [0.163, 1.116] & 0.962 / 0.040 \\
C-AdamW & 3.932 & 0.313 / 0.417 & 1.177 [0.596, 1.524] & 1.190 / 0.088 \\
SPAM & \textbf{3.884} & \textbf{0.187 / 0.269} & 0.236 [0.024, 0.839] & 1.045 / 0.594 \\
\bottomrule
\end{tabular}
\end{center}
\end{table*}

Table~\ref{tab:ftle_correlation} repeats the correlation analysis by channel over ten checkpoints. Final update separation correlates between 0.49 and 0.65 across channels, while the early log slope drops from 0.62 for weights to 0.10 for second moments: that response reaches the endpoint without proportional early growth. Batch order shows the weakest association, consistent with an external forcing term.

\begin{table*}[t]
\caption{Checkpoint-level Pearson correlations by perturbation channel.}
\label{tab:ftle_correlation}
\begin{center}
\small
\begin{tabular}{lrrr}
\toprule
Perturbation & Final update & Log-update slope & Early log slope \\
\midrule
$\theta$ & 0.652 & 0.719 & 0.624 \\
Adam-$m$ & 0.652 & 0.721 & 0.469 \\
Adam-$v$ & 0.615 & 0.424 & 0.095 \\
Batch order & 0.488 & 0.173 & -0.143 \\
\bottomrule
\end{tabular}
\end{center}
\end{table*}

\subsection{Scale transfer}

\begin{figure*}[t]
\begin{center}
\includegraphics[width=\linewidth]{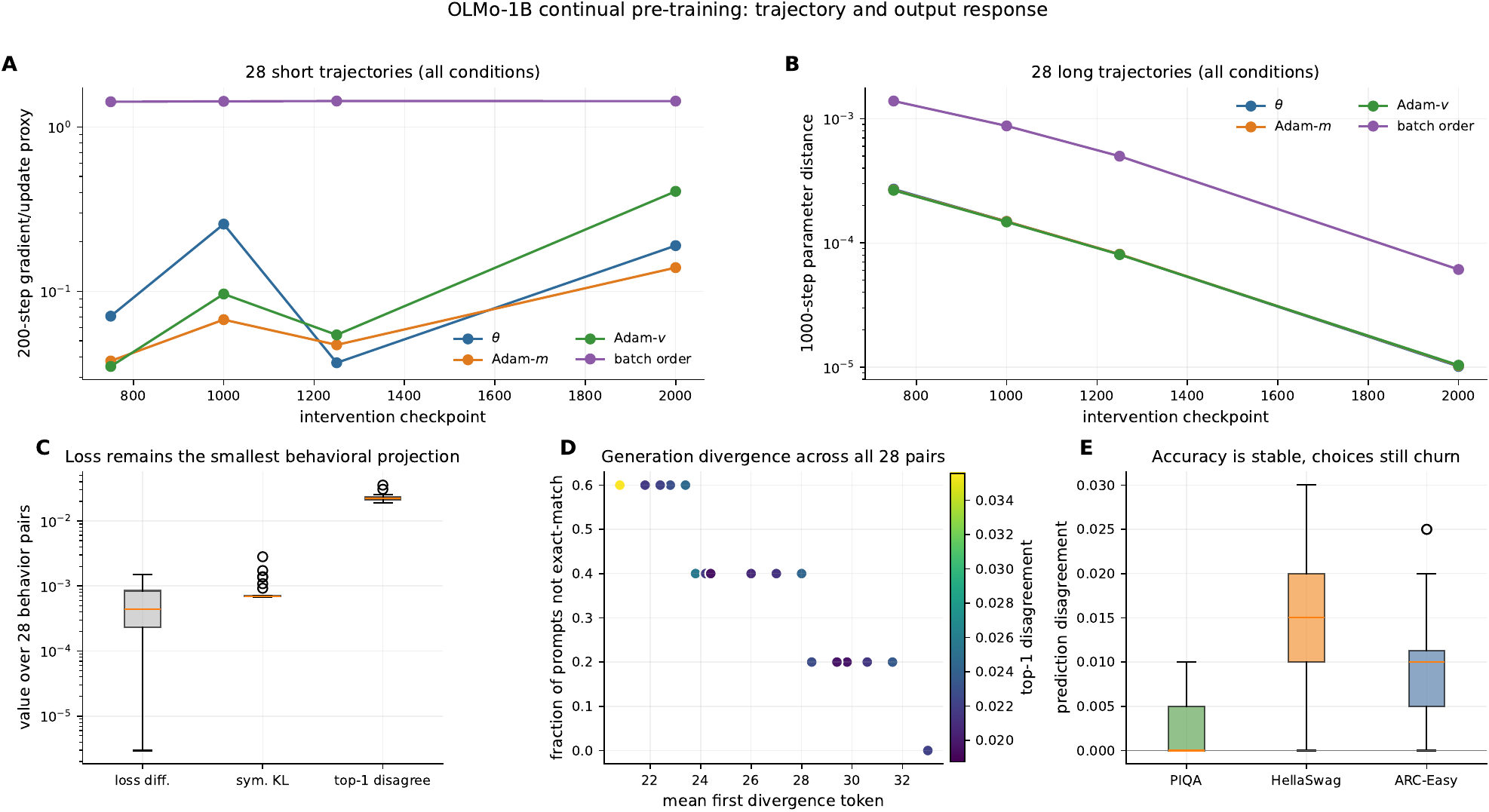}
\end{center}
\caption{OLMo-1B sensitivity. A--B: trajectory separation; C--E: output comparisons.}
\label{fig:olmo_transfer}
\end{figure*}

Figure~\ref{fig:olmo_transfer} gives the channel-level separations and the output comparisons behind Figure~\ref{fig:transfer}C--D. Continuations from the two selected checkpoints reproduce the small-model ordering at 1.485B parameters. Panels C--E report the token, generation, and benchmark differences at the two selected checkpoints. The same loss--output discrepancy appears at the larger scale: mean loss differences stay near $10^{-4}$ while prediction and generation differences remain nonzero.

\begin{figure*}[t]
\begin{center}
\includegraphics[width=0.82\linewidth]{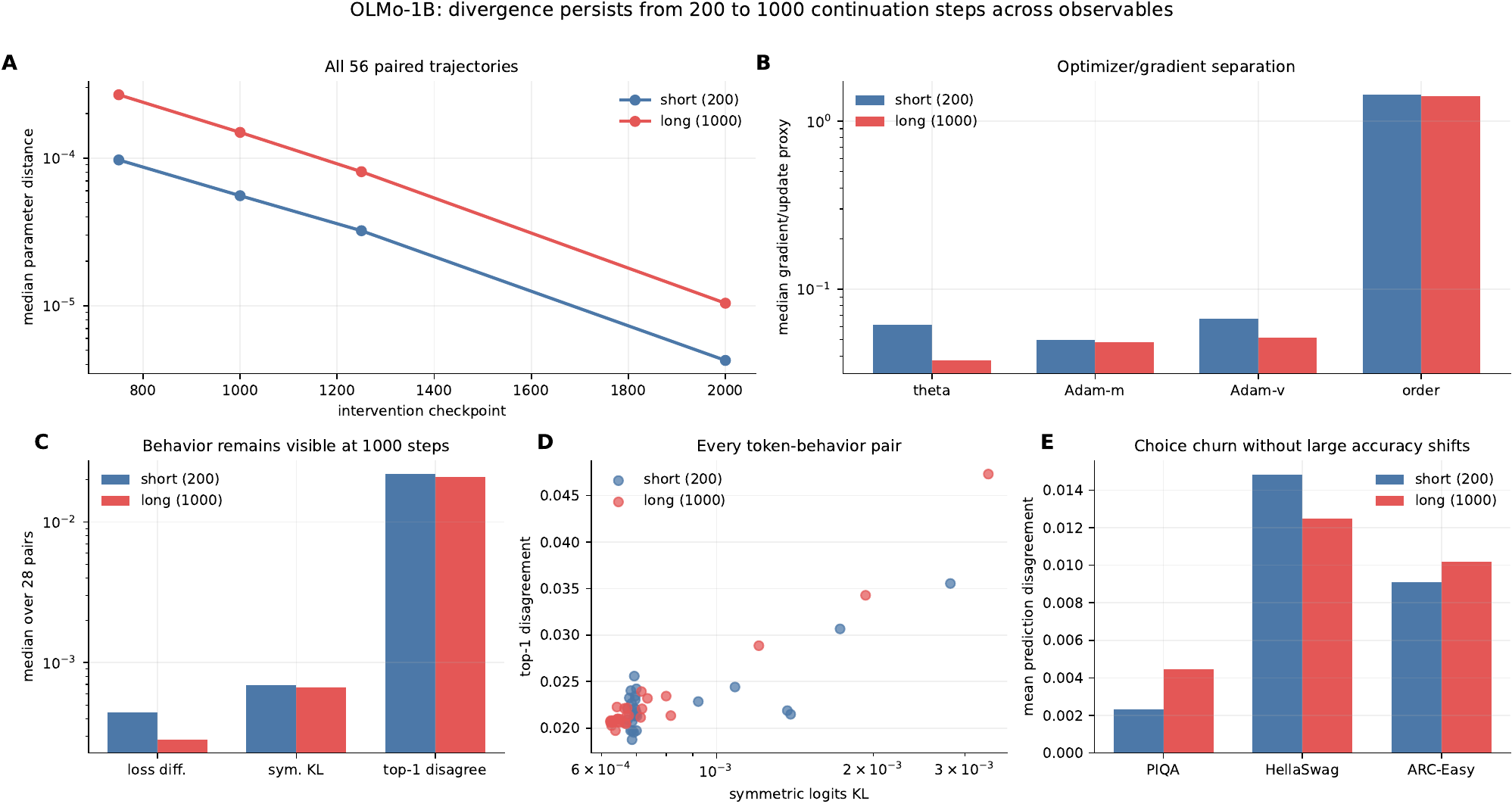}
\end{center}
\caption{OLMo-1B horizon comparison. Panels pair 200- and 1000-step outcomes.}
\label{fig:olmo_horizon}
\end{figure*}

Figure~\ref{fig:olmo_horizon} pairs the 200- and 1000-step outcome for each condition. Several channels contract in parameter and continuous logit distance between the two horizons, while top-1 disagreement and generation differences remain. The two horizons therefore separate transient trajectory growth from persistent behavioral difference, as the main text reports.

\subsection{Replication and behavior summaries}

\begin{figure*}[t]
\begin{center}
\includegraphics[width=\linewidth]{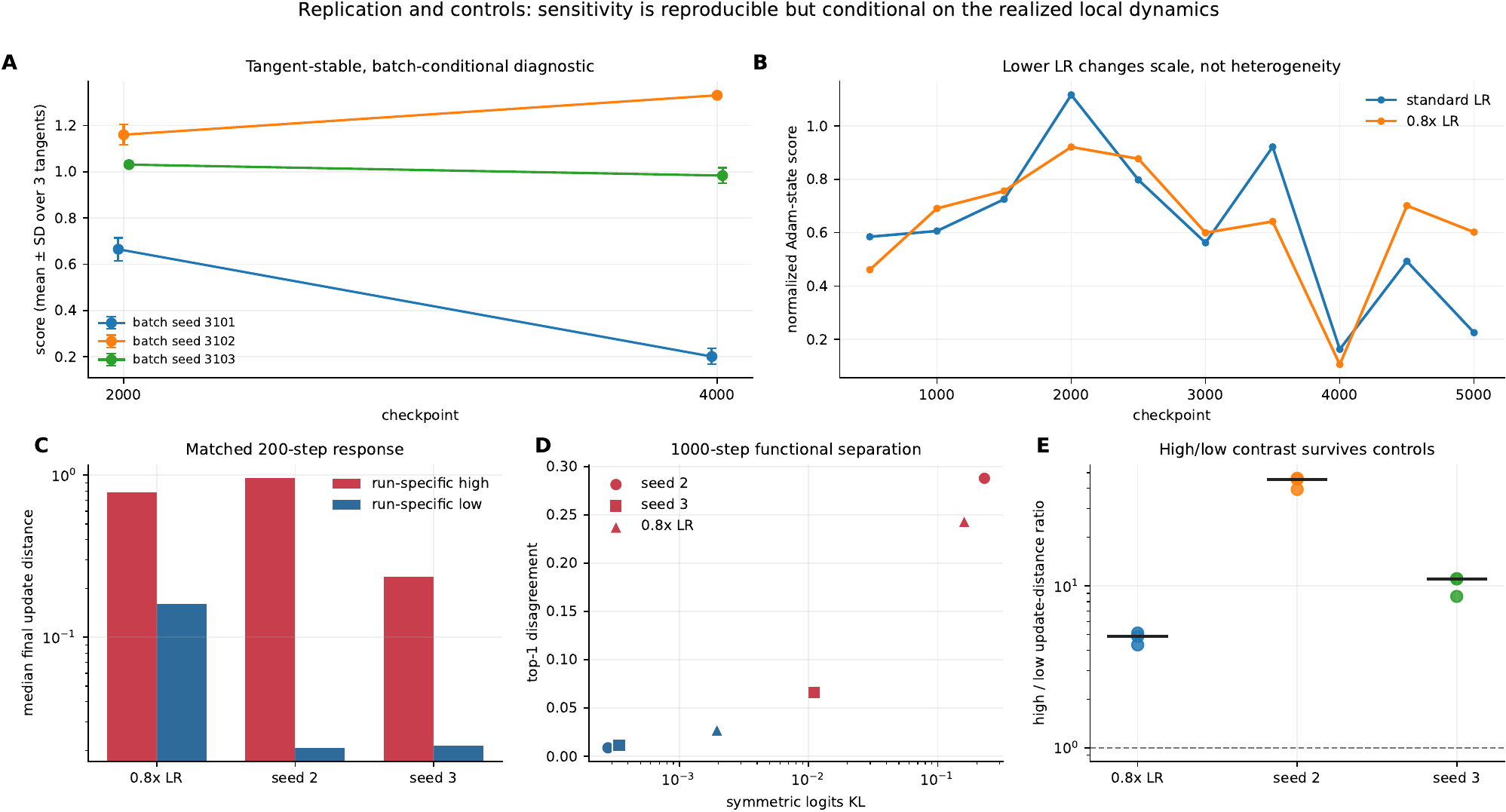}
\end{center}
\caption{Replication and diagnostic controls. Panels compare diagnostic batches, learning rate, and continuation behavior.}
\label{fig:replication_full}
\end{figure*}

Figure~\ref{fig:replication_full} collects the diagnostic-batch and learning-rate controls behind the seed-replication paragraph. The $0.8\times$-learning-rate run keeps a 0.783/0.160 high/low update-distance contrast. Repeated tangent seeds change the score less than alternative diagnostic batches, so a checkpoint value is conditional on the batch used to probe it.

\begin{figure*}[t]
\begin{center}
\includegraphics[width=\linewidth]{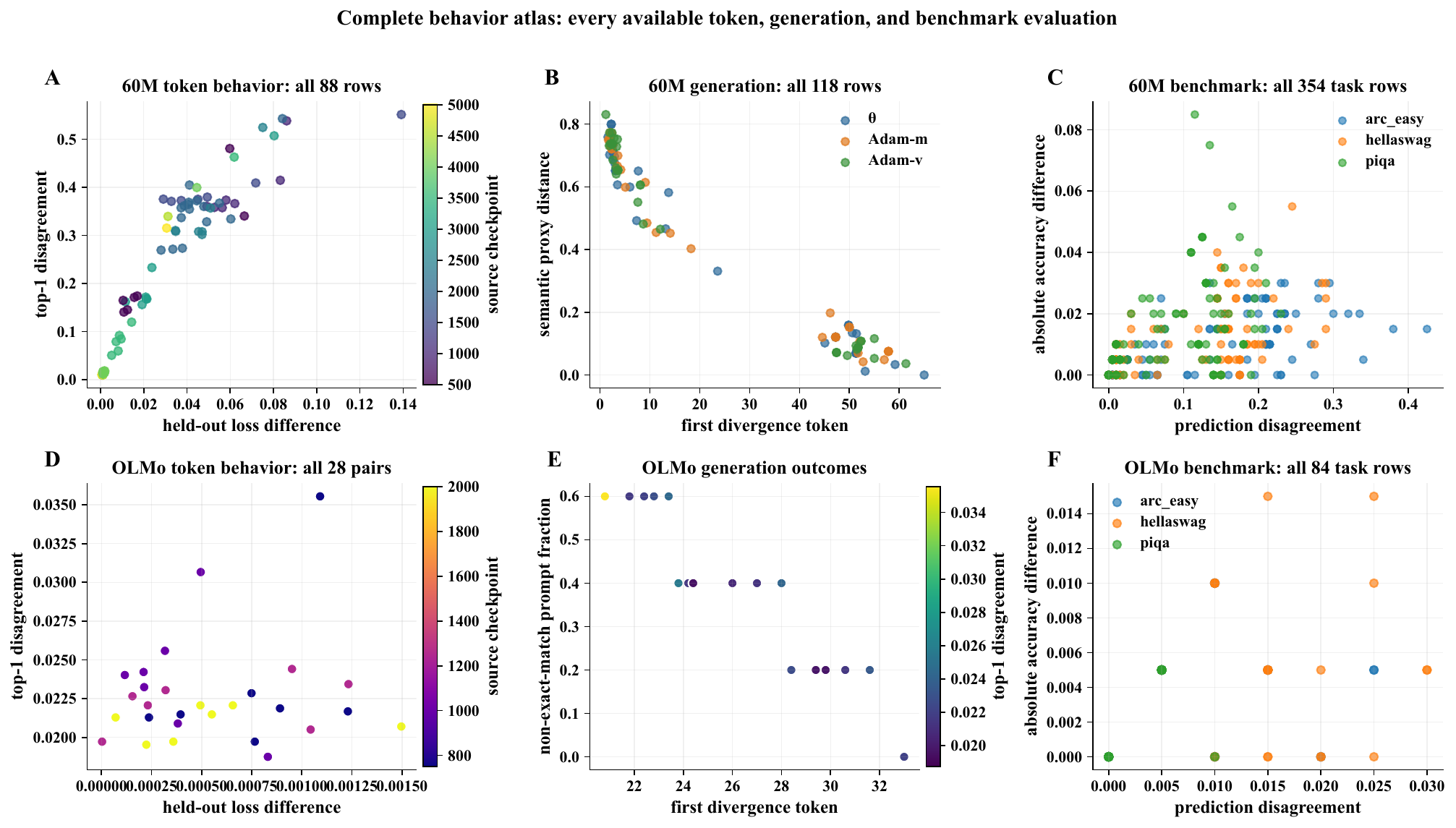}
\end{center}
\caption{Behavior across models. A--C: small-model token, generation, and benchmark outcomes; D--F: OLMo-1B.}
\label{fig:behavior_atlas}
\end{figure*}

Figure~\ref{fig:behavior_atlas} collects token, generation, and benchmark outcomes for the selected checkpoints in both systems, placing top-1 disagreement, generation match, and benchmark changes against trajectory separation. Token and benchmark changes increase with separation while net accuracy moves little relative to the number of changed answers, and generation agreement follows the same ordering.

The response ordering is stable across amplitude, horizon, optimizer, and scale; absolute magnitudes depend on the condition.

\section{Muon/AdamW Hybrid Pilot}
\label{app:muon}

\begin{figure*}[!t]
\begin{center}
\includegraphics[width=\linewidth]{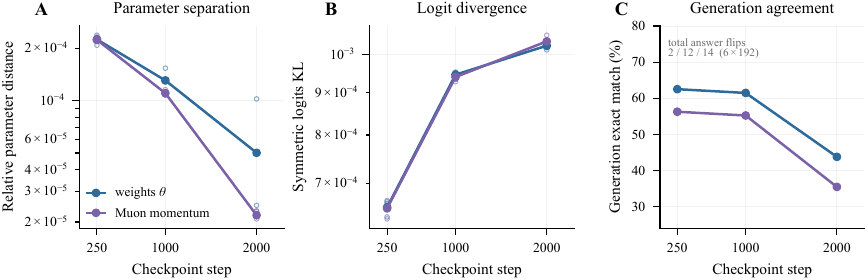}
\end{center}
\caption{Muon/AdamW hybrid response. A: relative parameter separation; B: symmetric logits KL; C: generation exact match, with the total number of changed benchmark answers across six conditions above each checkpoint (192 questions per condition). Lines average six conditions per checkpoint; open circles mark individual conditions.}
\label{fig:muon_pilot}
\end{figure*}

The pilot tests whether the separation--behavior gap depends on AdamW's second-moment preconditioner; Figure~\ref{fig:muon_pilot} summarizes the averaged trend and Tables~\ref{tab:muon_grid}--\ref{tab:muon_diag} list every condition and the finite-difference check.

The pilot trains OLMo-1B (1.485B parameters) for 2000 steps with Muon on the two-dimensional hidden matrices (momentum 0.95, Nesterov, five Newton--Schulz steps, AdamW-matched learning rate) and AdamW on the embeddings, LM head, and vector parameters. Optimizer states are stored in FP32 and passes run under BF16 autocast, so a relative perturbation of $10^{-6}$ survives the update. Each response loads the same checkpoint in two branches, perturbs the selected state, and advances 200 steps on shared batches under the global schedule. Outputs are measured on four fixed validation batches (4096 positions), 32 prompts of 64 greedy tokens, and 64 questions each from PIQA, HellaSwag, and ARC-Easy. The pilot contains 33 responses, six diagnostics, and a zero-perturbation control that returns identical branches.

Table~\ref{tab:muon_grid} shows that parameter separation contracts sixfold between steps 250 and 2000, symmetric KL grows by 1.6, and generation agreement drops from 59.4\% to 39.6\% (Figure~\ref{fig:muon_pilot}); top-1 disagreement stays between 1.1\% and 1.6\%, so the token-level measurement saturates before generation does. Across the nine matched amplitudes, the median momentum-to-weight ratio is 0.97 for update separation, 0.93 for parameter separation, and 0.97 for top-1 disagreement; parameter ratios span 0.20 to 1.05.

Table~\ref{tab:muon_states} collects the layer, optimizer-state, and data-stream conditions. Layer scope matters little: at step 1000 the six groups vary by less than a factor of 1.3 in parameter separation, with no depth ordering. The second moment raises parameter separation from $4.1\times10^{-4}$ to $1.8\times10^{-3}$ between $\epsilon=10^{-6}$ and $10^{-4}$, while the first moment stays near $1.1\times10^{-4}$, the same asymmetry as in the small-model and OLMo runs. The data-stream change keeps update separation near 1.39 at every checkpoint while parameter separation falls from $6.2\times10^{-3}$ to $6.8\times10^{-4}$ and token disagreement from 13.3\% to 3.0\%. Answer flips per condition are 0--4 of 192 (state) and 3--7 (data stream); no condition moves net accuracy by more than 4 of 192.

Table~\ref{tab:muon_diag} reports the three-step finite-difference growth $g_3=\frac13\log(N_3/N_0)$: values lie between 3.36 and 3.73 per step, and the two directions at step 1000 differ by 0.25, comparable to the spread across checkpoints. The Muon/AdamW state therefore amplifies a local perturbation within a few steps.

\begin{table*}[!t]
\caption{All 18 all-model Muon/AdamW conditions after 200 continuation steps. Top-1 and generation match are percentages over 4096 positions and 32 prompts; flips count changed answers per condition over the same 192 benchmark questions.}
\label{tab:muon_grid}
\begin{center}
\small
\setlength{\tabcolsep}{5pt}
\begin{tabular}{llcccccccc}
\toprule
ID & Step & Channel & $\epsilon$ & Update sep. & Parameter sep. & sKL & Top-1 & Gen. match & Flips\\
\midrule
M04 & 250 & weights & $10^{-6}$ & 0.070 & $2.29\times10^{-4}$ & $6.62\times10^{-4}$ & 1.20\% & 56.2\% & 0\\
M05 & 250 & weights & $10^{-5}$ & 0.062 & $2.08\times10^{-4}$ & $6.37\times10^{-4}$ & 1.44\% & 65.6\% & 0\\
M06 & 250 & weights & $10^{-4}$ & 0.064 & $2.38\times10^{-4}$ & $6.65\times10^{-4}$ & 1.22\% & 65.6\% & 1\\
M07 & 250 & Muon mom. & $10^{-6}$ & 0.067 & $2.32\times10^{-4}$ & $6.67\times10^{-4}$ & 1.12\% & 65.6\% & 0\\
M08 & 250 & Muon mom. & $10^{-5}$ & 0.063 & $2.19\times10^{-4}$ & $6.57\times10^{-4}$ & 1.39\% & 53.1\% & 1\\
M09 & 250 & Muon mom. & $10^{-4}$ & 0.066 & $2.19\times10^{-4}$ & $6.34\times10^{-4}$ & 1.17\% & 50.0\% & 0\\
M13 & 1000 & weights & $10^{-6}$ & 0.068 & $1.10\times10^{-4}$ & $9.51\times10^{-4}$ & 1.56\% & 59.4\% & 2\\
M14 & 1000 & weights & $10^{-5}$ & 0.061 & $1.28\times10^{-4}$ & $9.34\times10^{-4}$ & 1.54\% & 68.8\% & 3\\
M15 & 1000 & weights & $10^{-4}$ & 0.062 & $1.54\times10^{-4}$ & $9.53\times10^{-4}$ & 1.64\% & 56.2\% & 3\\
M16 & 1000 & Muon mom. & $10^{-6}$ & 0.062 & $1.08\times10^{-4}$ & $9.44\times10^{-4}$ & 1.51\% & 56.2\% & 1\\
M17 & 1000 & Muon mom. & $10^{-5}$ & 0.069 & $1.16\times10^{-4}$ & $9.48\times10^{-4}$ & 1.59\% & 56.2\% & 2\\
M18 & 1000 & Muon mom. & $10^{-4}$ & 0.066 & $1.08\times10^{-4}$ & $9.28\times10^{-4}$ & 1.51\% & 53.1\% & 1\\
M22 & 2000 & weights & $10^{-6}$ & 0.076 & $2.50\times10^{-5}$ & $1.03\times10^{-3}$ & 1.32\% & 37.5\% & 2\\
M23 & 2000 & weights & $10^{-5}$ & 0.086 & $2.30\times10^{-5}$ & $1.01\times10^{-3}$ & 1.10\% & 46.9\% & 2\\
M24 & 2000 & weights & $10^{-4}$ & 0.068 & $1.02\times10^{-4}$ & $1.04\times10^{-3}$ & 1.39\% & 46.9\% & 2\\
M25 & 2000 & Muon mom. & $10^{-6}$ & 0.058 & $2.33\times10^{-5}$ & $1.04\times10^{-3}$ & 1.34\% & 34.4\% & 2\\
M26 & 2000 & Muon mom. & $10^{-5}$ & 0.063 & $2.16\times10^{-5}$ & $1.02\times10^{-3}$ & 1.32\% & 43.8\% & 2\\
M27 & 2000 & Muon mom. & $10^{-4}$ & 0.062 & $2.09\times10^{-5}$ & $1.06\times10^{-3}$ & 1.27\% & 28.1\% & 4\\
\bottomrule
\end{tabular}
\end{center}
\end{table*}

\begin{table*}[!t]
\caption{Layer, optimizer-state, and data-stream conditions after 200 continuation steps.}
\label{tab:muon_states}
\begin{center}
\small
\setlength{\tabcolsep}{7pt}
\begin{tabular}{llllcccccc}
\toprule
ID & Step & Channel & Scope & $\epsilon$ & Update sep. & Parameter sep. & sKL & Top-1 & Gen. match\\
\midrule
M29 & 1000 & weights & shallow & $10^{-5}$ & 0.070 & $1.43\times10^{-4}$ & $9.71\times10^{-4}$ & 1.73\% & 50.0\%\\
M30 & 1000 & weights & middle & $10^{-5}$ & 0.084 & $1.09\times10^{-4}$ & $9.27\times10^{-4}$ & 1.46\% & 59.4\%\\
M31 & 1000 & weights & deep & $10^{-5}$ & 0.067 & $1.24\times10^{-4}$ & $9.54\times10^{-4}$ & 1.54\% & 53.1\%\\
M32 & 1000 & Muon mom. & shallow & $10^{-5}$ & 0.069 & $1.12\times10^{-4}$ & $9.32\times10^{-4}$ & 1.59\% & 62.5\%\\
M33 & 1000 & Muon mom. & middle & $10^{-5}$ & 0.062 & $1.09\times10^{-4}$ & $9.16\times10^{-4}$ & 1.37\% & 50.0\%\\
M34 & 1000 & Muon mom. & deep & $10^{-5}$ & 0.062 & $1.10\times10^{-4}$ & $9.20\times10^{-4}$ & 1.56\% & 68.8\%\\
M35 & 1000 & Adam-$m$ & all & $10^{-6}$ & 0.064 & $1.07\times10^{-4}$ & $9.20\times10^{-4}$ & 1.61\% & 59.4\%\\
M36 & 1000 & Adam-$m$ & all & $10^{-5}$ & 0.070 & $1.19\times10^{-4}$ & $9.57\times10^{-4}$ & 1.59\% & 59.4\%\\
M37 & 1000 & Adam-$m$ & all & $10^{-4}$ & 0.068 & $1.16\times10^{-4}$ & $9.49\times10^{-4}$ & 1.49\% & 75.0\%\\
M38 & 1000 & Adam-$v$ & all & $10^{-6}$ & 0.064 & $4.14\times10^{-4}$ & $9.44\times10^{-4}$ & 1.83\% & 56.2\%\\
M39 & 1000 & Adam-$v$ & all & $10^{-5}$ & 0.071 & $7.06\times10^{-4}$ & $9.47\times10^{-4}$ & 1.71\% & 43.8\%\\
M40 & 1000 & Adam-$v$ & all & $10^{-4}$ & 0.077 & $1.78\times10^{-3}$ & $1.11\times10^{-3}$ & 1.86\% & 53.1\%\\
M10 & 250 & data stream & all & -- & 1.391 & $6.24\times10^{-3}$ & $4.88\times10^{-2}$ & 13.26\% & 3.1\%\\
M19 & 1000 & data stream & all & -- & 1.397 & $3.37\times10^{-3}$ & $2.24\times10^{-2}$ & 8.69\% & 3.1\%\\
M28 & 2000 & data stream & all & -- & 1.396 & $6.77\times10^{-4}$ & $3.71\times10^{-3}$ & 2.98\% & 12.5\%\\
\bottomrule
\end{tabular}
\end{center}
\end{table*}

\begin{table*}[!t]
\caption{Three-step finite-difference state growth for weight perturbations at $\epsilon=10^{-6}$.}
\label{tab:muon_diag}
\begin{center}
\small
\begin{tabular}{lrcrrr}\toprule
ID & Step & Direction & $N_0$ & $N_3$ & $g_3$\\\midrule
M02 & 250 & 9101 & $1.00\times10^{-6}$ & 0.0262 & 3.391\\
M03 & 250 & 9102 & $1.00\times10^{-6}$ & 0.0242 & 3.364\\
M11 & 1000 & 9101 & $1.00\times10^{-6}$ & 0.0715 & 3.726\\
M12 & 1000 & 9102 & $1.00\times10^{-6}$ & 0.0339 & 3.477\\
M20 & 2000 & 9101 & $1.00\times10^{-6}$ & 0.0265 & 3.395\\
M21 & 2000 & 9102 & $1.00\times10^{-6}$ & 0.0264 & 3.393\\\bottomrule
\end{tabular}
\end{center}
\end{table*}

\end{document}